\documentclass[10pt,twocolumn,letterpaper]{article}

\usepackage{cvpr}              
\usepackage{times}
\usepackage{color,xcolor}
\usepackage{epsfig}
\usepackage{graphicx}

\usepackage{adjustbox}
\usepackage{array}
\usepackage{booktabs}
\usepackage{colortbl}
\usepackage{float,wrapfig}
\usepackage{hhline}
\usepackage{multirow}
\usepackage{caption}
\usepackage{amsmath,amsfonts,amsthm,amssymb}
\usepackage{bm}
\usepackage{nicefrac}
\usepackage{microtype}

\usepackage{changepage}
\usepackage{extramarks}
\usepackage{fancyhdr}
\usepackage{lastpage}
\usepackage{setspace}
\usepackage{soul}
\usepackage{xspace}
\usepackage{dsfont}

\usepackage{algorithm}
\usepackage{algpseudocode}
\usepackage{enumerate}
\usepackage{comment}

\usepackage{url}

\definecolor{cvprblue}{rgb}{0.21,0.49,0.74}

\definecolor{nvgreen}{RGB}{118, 185, 0}

\usepackage[breaklinks,colorlinks,urlcolor=nvgreen,citecolor=cvprblue,linkcolor=cvprblue]{hyperref}

\usepackage{pifont}
\usepackage{nccmath}

\usepackage{soul}
\sethlcolor{red}

\usepackage{paralist}

\newcommand{\myparagraph}[1]{\vspace{-5pt}\paragraph{#1}}

\newcommand{\ignorethis}[1]{}

\renewcommand*{\thefootnote}{\fnsymbol{footnote}}

\makeatletter
\DeclareRobustCommand\onedot{\futurelet\@let@token\@onedot}
\def\@onedot{\ifx\@let@token.\else.\null\fi\xspace}

\makeatother

\makeatletter
\newcommand\footnoteref[1]{\protected@xdef\@thefnmark{\ref{#1}}\@footnotemark}
\makeatother

\definecolor{mydarkblue}{rgb}{0,0.08,1}
\definecolor{mydarkred}{rgb}{0.8,0.02,0.02}
\definecolor{mydarkorange}{rgb}{0.40,0.2,0.02}
\definecolor{mypurple}{RGB}{111,0,255}
\definecolor{myred}{rgb}{1.0,0.0,0.0}
\definecolor{mygold}{rgb}{0.75,0.6,0.12}
\definecolor{mydarkgray}{rgb}{0.66, 0.66, 0.66}
\definecolor{mydarkgreen}{rgb}{0.02,0.6,0.02}
\definecolor{mygray}{gray}{0.9}
\definecolor{keynotegreen}{rgb}{0.04,0.52,0}
\definecolor{keynoteyellow}{rgb}{1,0.68,0}

\newcommand\blfootnote[1]{%
	\begingroup
	\renewcommand\thefootnote{}\footnote{#1}%
	\addtocounter{footnote}{-1}%
	\endgroup
}

\newcommand{\invalid}[1]{{\color{gray!50}{}#1{}}}

\def\methodshort{VILA\xspace}

\def\paperID{5738}
\def\confName{CVPR}
\def\confYear{2024}

\begin{document}

\title{\methodshort: On Pre-training for Visual Language Models}

\author{Ji Lin\textsuperscript{1,2 * \dag} 
\qquad
Hongxu Yin\textsuperscript{1 *}
\qquad
Wei Ping\textsuperscript{1}
\qquad
Yao Lu\textsuperscript{1} 
\qquad
Pavlo Molchanov\textsuperscript{1}
\qquad\\
Andrew Tao \textsuperscript{1} 
\qquad
Huizi Mao \textsuperscript{1} 
\qquad
Jan Kautz\textsuperscript{1} 
\qquad
Mohammad Shoeybi\textsuperscript{1} 
\qquad
Song Han\textsuperscript{1,2}
\\
$^1$NVIDIA \quad $^2$MIT \quad \\
\small{\includegraphics[height=.6\baselineskip]{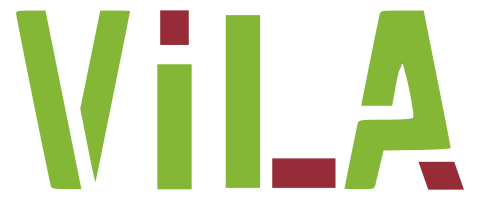} \ \ \url{https://github.com/NVlabs/VILA}} \\
}

\maketitle


\begin{abstract}
\blfootnote{$^*$ Equal contribution. $^\dag$ Work done during an internship at NVIDIA.}

Visual language models (VLMs) rapidly progressed with the recent success of large language models. 
There have been growing efforts on visual \emph{instruction tuning} to extend the LLM with visual inputs, but lacks an in-depth study of the visual language \emph{pre-training} process, where the model learns to perform joint modeling on both modalities. 
In this work, we examine the design options for VLM pre-training by augmenting LLM towards VLM through step-by-step controllable comparisons. We introduce three main findings: (1) freezing LLMs during pre-training can achieve decent zero-shot performance, but lack in-context learning capability, which requires unfreezing the LLM; (2) interleaved pre-training data is beneficial whereas image-text pairs alone are not optimal; (3) re-blending text-only instruction data to image-text data during instruction fine-tuning not only remedies the degradation of text-only tasks, but also boosts VLM task accuracy. 
With an enhanced pre-training recipe we build \textbf{VILA},  a \textbf{Vi}sual \textbf{La}nguage model family that 
consistently outperforms the state-of-the-art models, \textit{e.g.}, LLaVA-1.5, 
across main benchmarks without bells and whistles.
Multi-modal pre-training also helps unveil appealing properties of VILA, including multi-image reasoning, enhanced in-context learning, and better world knowledge. VILA is also \href{https://github.com/mit-han-lab/llm-awq/tree/main/tinychat}{deployable} on Jetson Orin for on-device VLM.

\end{abstract}

\section{Introduction}

\begin{figure}[t]
    \centering
     \includegraphics[width=0.45\textwidth]{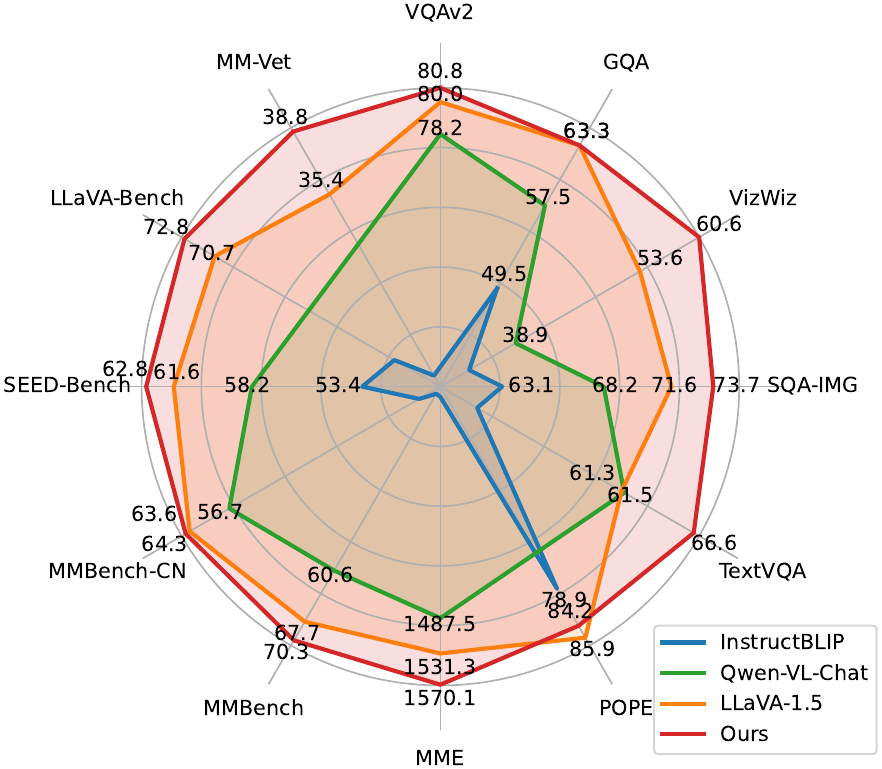}
    \caption{VILA's enhanced visual-language pre-training consistently improves the downstream task accuracy under a 
    comparison to recent methods~\cite{qwen,Dai2023InstructBLIP,liu2023llava}. 
    } 
    \label{fig:teaser}
\end{figure}

Large language models (LLMs) have demonstrated superior capabilities for natural language tasks~\cite{raffel2020exploring,dai2019transformer, gpt3, OpenAI_ChatGPT,touvron2023llama, touvron2023llama2,alpaca,vicuna2023,karamcheti2021mistral,chowdhery2022palm,yi,qwen}.
Augmenting LLMs to support visual inputs allows the final model to inherit some of the appealing properties like instruction following, zero-shot generalization, and few-shot in-context learning (ICL), empowering various visual language tasks~\cite{liu2023llava,alayrac2022flamingo,driess2023palm,chen2023pali,li2023blip,fuyu,bai2023qwen,GPT4,zhu2023minigpt}. 
The central challenge of unifying vision and language for collaborative inference resides in connecting the LLM and the vision foundation model (\eg, a CLIP encoder): both foundation models are usually pre-trained individually, before aligned via vision-language joint training. 
Most of the efforts in this field have been focusing on improving the visual language instruction-tuning process, \ie, supervised fine-tuning (SFT) or reinforcement learning from
human feedback (RLHF)~\cite{liu2023llava, liu2023improved, sun2023aligning}. However, there lacks a thorough study of the pre-training process, where the model is trained on image-text datasets/corpora at scale~\cite{zhu2023multimodal,kakaobrain2022coyo-700m,schuhmann2022laion}. This process is costly but critical for the modality alignment.


In this work, we aim to explore different design options for enhanced visual language model pre-training. 
In particular, we aim to answer \textit{``How do various design choices in visual language model pre-training impact the downstream performance?"}
We followed the pre-training + SFT pipeline and ablated different design options for pre-training overseeing dataset properties and training protocols. 
We discover several findings: 
(1) Freezing the LLM during pre-training can achieve a decent zero-shot performance, but not in-context learning~(ICL) capability, whereas updating the LLMs encourages deep embedding alignment, which we found is important for ICL; 
(2) Interleaved visual language data is essential for pre-training, that provides accurate gradient update and maintains text-only capability; 
(3) Adding in text-only instruction data during SFT can further remedy text-only degradation and boost visual language task accuracy. 

We introduce practical guidance to design \textbf{Vi}sual \textbf{La}nguage models, dubbed \textbf{VILA}. 
Without bells and whistles, \methodshort outperforms the state-of-the-art model~\cite{liu2023improved} by noticeable margins across a wide range of vision language tasks (Figure~\ref{fig:teaser}), thanks to the help of improved pre-training. Moreover, we observe that the pre-training process unlocked several interesting capabilities for the model, such as (i) multi-image reasoning (despite the model only sees single image-text pairs during SFT), (ii) stronger in-context learning capabilities, and (iii) enhanced world knowledge.
We hope our findings can provide a good pre-training recipe for future visual language models. 


\section{Background}

\begin{figure}[t]
    \centering
     \includegraphics[width=0.5\textwidth]{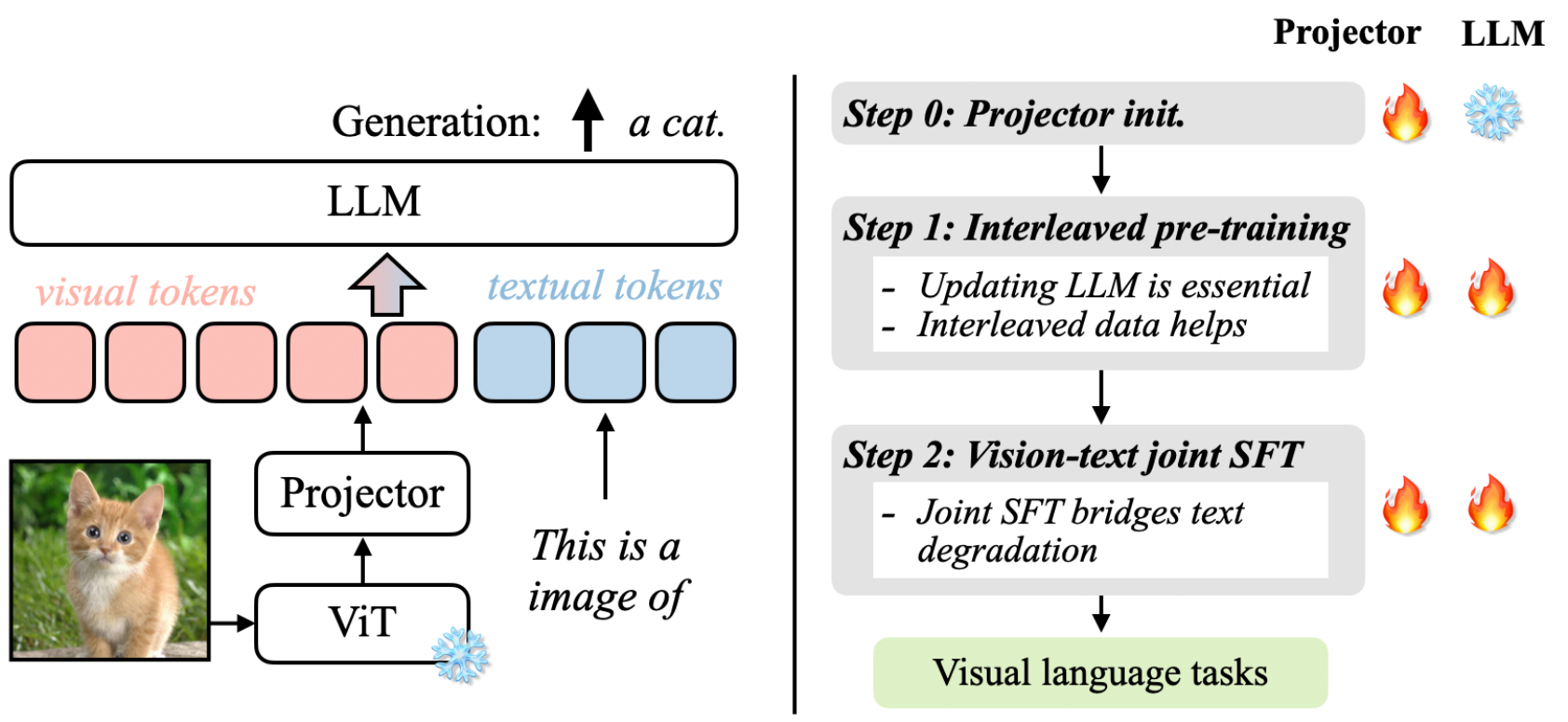}
    \caption{We study auto-regressive visual language model, where images are tokenized and fed to the input of LLMs. We find updating the LLM is essential for in-context learning capabilities, and interleaved corpus like~\cite{zhu2023multimodal} helps pre-training. Joint SFT with text-only data helps maintain the text-only capabilities. 
    } 
    \label{fig:arch}
\end{figure}

\myparagraph{Model architecture. }
Multi-modal LLMs can be generally categorized into two settings: 
cross-attention-based~\citep{alayrac2022flamingo,li2023blip} and auto-regressive-based~\citep{driess2023palm, liu2023llava, fuyu}. 
The latter VLM family tokenizes images into visual tokens, which are concatenated with textual tokens and fed as the input to LLMs (\ie, treating visual input as a foreign language). 
It is a natural extension of text-only LLMs by augmenting the input with visual embeddings and can handle arbitrary interleaved image-text inputs.
In this study, we focus on the pre-training of auto-regressive  VLMs due to its flexibility and popularity. As shown in Figure~\ref{fig:arch}, auto-regressive VLMs consists of three components: a \emph{visual encoder},
an \emph{LLM}, and a \emph{projector} that bridges the embeddings from the two modalities. The projector can be a simple linear layer~\cite{liu2023llava} or more capable Transformer blocks~\cite{openflamingo, Dai2023InstructBLIP} -- we will compare their efficacy in our experiments. 
The model takes visual and text input and generates text outputs.


\myparagraph{Training stages. } 
Following common practice~\cite{openflamingo, driess2023palm, liu2023llava}, we study how to augment a pre-trained text-only LLM with visual input support. The training can be categorized into three stages: 

\emph{0. Projector initialization}. The LLM and ViT are separately pre-trained, while the projector is usually initialized from random weights. Therefore, 
we first pre-train the projector while freezing both ViT and LLMs on image-caption pairs following existing literature~\citep{liu2023llava, li2023blip, Dai2023InstructBLIP}.

\emph{1. visual language pre-training}. We then pre-train the model (LLM and the projector) on visual language corpus.
We consider two types of corpus: interleaved image-text corpus (\eg, MMC4~\citep{zhu2023multimodal}) and image-text pairs (\eg, COYO~\cite{kakaobrain2022coyo-700m} and LAION~\cite{schuhmann2022laion}). 
We focus the study of this work on the pre-training process, which are most costly and important for visual language alignment. 

\emph{2. Visual instruction-tuning}. Finally, we further perform instruction tuning of the pre-trained model on visual language instruction datasets.
We convert existing visual language datasets into FLAN~\citep{wei2021finetuned} style (\ie, with dataset-specific prompts) following~\citep{Dai2023InstructBLIP}. 
Please find the data blend of the visual instruction data in the supplementary.

\myparagraph{Evaluations.}
During our ablation study, we evaluate the fine-tuned model on 4 visual language tasks: accuracy for OKVQA~\cite{marino2019okvqa} and TextVQA~\cite{singh2019textvqa}, and CIDEr score for COCO~\cite{lin2014mscoco} and Flickr~\cite{young2014flickr}. We evaluate both 0-shot and 4-shot performance, which reflects the models' in-context learning capability. 

    
    
    

\section{On Pre-training for Visual Language Models}

In this section, we discuss practical design choices and learned lessons for the visual language pre-training process. 
\subsection{Updating LLM is Essential}

\begin{table*}[t]
    \small
    \centering
    \begin{tabular}{lcclcccccccccc}
        \toprule
        &  PreT & SFT & \multirow{2}{*}{Projector} & \multicolumn{2}{c}{OKVQA} & \multicolumn{2}{c}{TextVQA} & \multicolumn{2}{c}{COCO} & \multicolumn{2}{c}{Flickr} &  \multicolumn{2}{c}{\textbf{Average}}     \\ \cmidrule(lr){2-3} \cmidrule(lr){5-6} \cmidrule(lr){7-8} \cmidrule(lr){9-10} \cmidrule(lr){11-12} \cmidrule(lr){13-14}
        &\multicolumn{2}{c}{Train LLM?} &  & 0-shot & 4-shot & 0-shot & 4-shot & 0-shot & 4-shot & 0-shot & 4-shot &0-shot & 4-shot \\ \midrule
         (a) & \ding{55} & \ding{55} & Transformer & 10.4 & 19.2 & 14.8 & 23.1 & 17.4 & 60.2 & 11.0 & 47.4 & \invalid{13.4} & \invalid{37.5} \\ \midrule
         (b) & \ding{55} & \ding{51} & Transformer & 47.1 & 47.7 & 37.2 & 36.6 & 109.4 & 88.0 & 73.6 & 58.1 & 66.8 & \invalid{57.6} \\
         (c) & \ding{51} & \ding{51} & Transformer & 44.8 & 49.8 & 38.5 & 38.8 & 112.3 & 113.5 & 71.5 & 72.9 & 66.8 & 68.8\\
         (d) & \ding{51} & \ding{51} & Linear & 45.2 & 50.3 & 39.7 & 40.2 & 115.7 & 118.5 & 74.2 & 74.7 & 68.7 & 70.9 \\ 
         
        \bottomrule
    \end{tabular}
    \caption{Ablation study on whether to train LLM or freeze LLM and only perform prompt tuning 
    during visual language pre-training (PreT). Interestingly, freezing the LLM during pre-training does not hurt the 0-shot accuracy, but leads to worse in-context learning capability (worse 4-shot). Using a simple linear projector forces the LLM to learn more and leads to better generalization. 
    We report accuracy for VQA datasets (OKVQA, TextVQA) and CIDEr score for captioning (COCO and Flickr). \emph{Note}: we used a different evaluation setting just for ablation study; the absolute value in this setting is lower and should not be compared against other work. }
    \label{tab:training_ablate}
\end{table*}

\begin{figure}[t]
    \centering
     \includegraphics[width=0.4\textwidth]{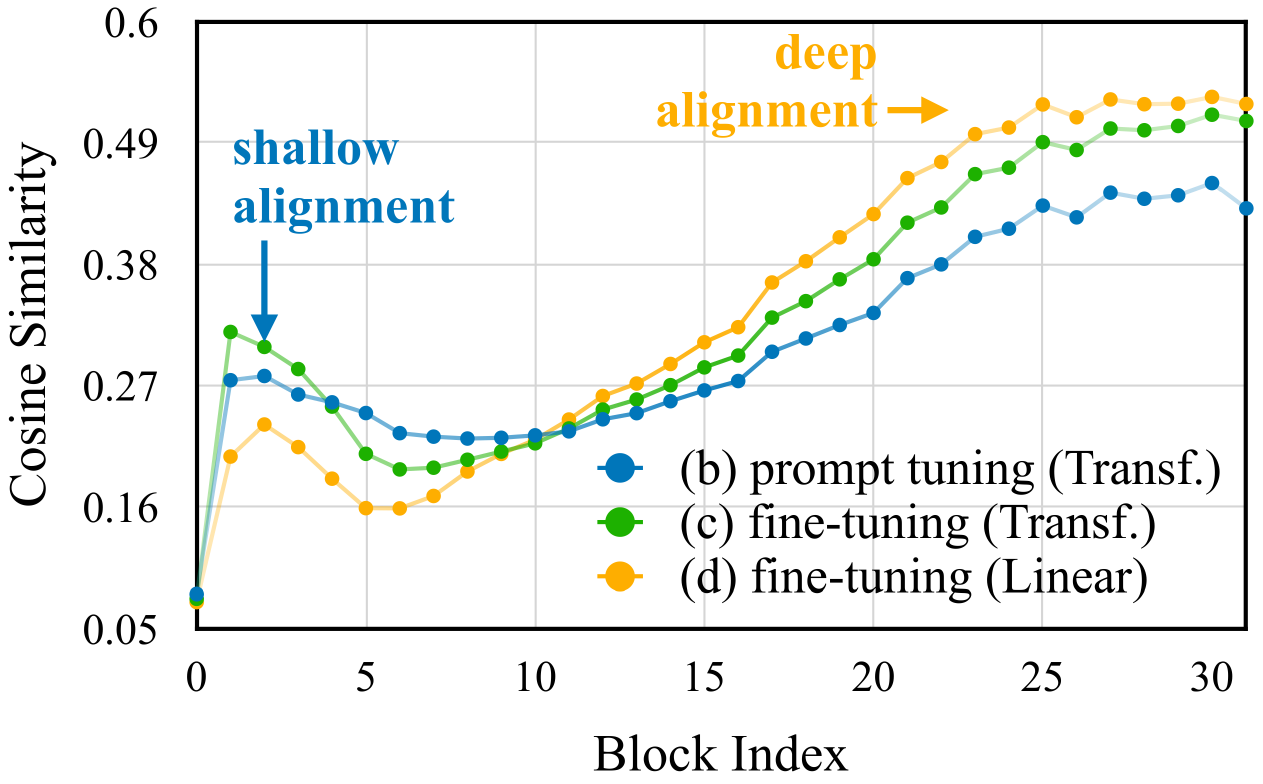}
    \caption{Prompt-tuning to support visual tokens can only enable shallow alignment, while fine-tuning the LLM leads to alignment at deeper layers. From configuration (b) to (d) (as in Table~\ref{tab:training_ablate}), the alignment improves at deeper layer, so as ICL accuracy (4-shot). 
    } 
    \label{fig:cos_sim}
\end{figure}

\myparagraph{Fine-tuning \vs prompt tuning. } 
There are two popular ways to augment a pre-trained text-only LM with visual inputs: \emph{fine-tune} LLMs on the visual input tokens~\cite{driess2023palm, liu2023llava}, or \emph{freeze} the LLM and train only the visual input projector as \emph{prompt tuning}~\citep{li2023blip, Dai2023InstructBLIP}. 
The latter is attractive since freezing the LLMs prevents the degradation of the pre-trained text-only LLM.  Nonetheless, we found updating the base LLM is essential to inherit some of the appealing LLM properties like in-context learning.

To verify the idea, we compare the two training protocols in Table~\ref{tab:training_ablate}. We use a Transformer block for the projector instead of a single linear layer~\cite{liu2023llava} in setting a-c, which provides enough capacity when freezing LLMs. We use MMC4-core~\cite{zhu2023multimodal}\footnote{We downloaded only $25$M of $30$M images amid some expired URLs. } for the comparison. We observed that:

(1) Training only the projector during SFT leads to poor performance (setting a), despite using a high-capacity design. It is rewarding to fine-tune LLM during SFT. 

(2) Interestingly, freezing the LLM during pre-training does \emph{not} affect \emph{0-shot performance}, but \emph{degrades in-context learning capabilities} (\ie, 4-shot, comparing setting b and c).
The gap is even larger for captioning datasets (COCO \& Flickr) since they are out-of-distribution (the instruction tuning data is mostly VQA-alike, see supplementary), showing the worse generalization capability when freezing LLMs. 

(3) When using a small-capacity projector (a linear layer instead of a Transformer block), the accuracy is slightly better (comparing c and d). We hypothesize a simpler projector forces the LLM to learn more on handling visual inputs, leading to better generalization.

\myparagraph{The deep embedding alignment hypothesis. }
To understand why fine-tuning LLM is beneficial, we hypothesize that it is important to \emph{align the distribution of visual and textual latent embeddings} (especially in the deeper layers), so that the model can seamlessly model the interaction between the two modalities. 
It is essential if we want to inherit some of the good properties of LLM like in-context learning for visual language applications.

To verify the idea, we calculate the Chamfer distance of visual and textual embeddings in different layers to measure how well they align in Figure~\ref{fig:cos_sim}. We calculate the pairwise cosine similarity to exclude the affect of magnitude. 
From configuration (b) to (d), the similarity of deeper layer goes higher, so as the 4-shot accuracy in Table~\ref{tab:training_ablate}, showing the positive relationship between deep embedding alignment and in-context learning. 

Given the observations, we \emph{fine-tune the LLM during both pre-training and instruction-tuning} in later studies, and use a \emph{simple linear projection} layer. 

\subsection{Interleaved Visual Language Corpus Helps Pre-training}
\label{sec:interleave_data}

\begin{table}[t]
    \setlength{\tabcolsep}{3.5pt}
    \small
    \centering
    \begin{tabular}{lcccc}
        \toprule
        Dataset & Type & Text Src. & \#img/sample & \#tok./img\\ \midrule
        MMC4~\cite{zhu2023multimodal} & Interleave & HTML & 4.0 & 122.5 \\ 
        COYO~\cite{kakaobrain2022coyo-700m} & Img-text pair & Alt-text & 1 & 22.7 \\
        \bottomrule
    \end{tabular}
    \caption{ Two image-text corpus considered for pre-training. The COYO captions are generally very short, which has a different distribution compared to the text-only corpus for LLM training.
    We sample each data source to contain 25M images by choosing samples with high CLIP similarities.
    }
    \label{tab:pretrain_data}
\end{table}

\begin{table}[t]
    \small
    \centering
    \begin{tabular}{lccl}
        \toprule
        \multirow{2}{*}{\shortstack{Pre-train\\Data}} &  \multicolumn{2}{c}{VLM acc (avg)} &  \multirow{2}{*}{MMLU acc.}    \\  \cmidrule(lr){2-3}  
         &  0-shot & 4-shot  \\ \midrule
        \rowcolor{black!10}
        \emph{Llama-2} & - & - & \emph{46.0\%} \\ 
        COYO &  51.1\% & 50.3\%  & 28.8\% (\textcolor{mydarkred}{-17.2\%}) \\
        MMC4-pairs & 46.4\% & 44.5\% & 32.4\% (\textcolor{mydarkred}{-13.6\%})\\
        MMC4 & 68.7\% & 70.9\%   & 40.7\%  (-5.3\%) \\ 
        \midrule
        MMC4+COYO  & \textbf{69.0\%} & \textbf{71.3\%} & 40.2\% (-5.8\%) \\ 
        \bottomrule
    \end{tabular}
    \caption{Pre-training on MMC4 data provides better visual language accuracy (0-shot and few-shot) and smaller degradation on text-only accuracy compared to caption data (COYO). The benefits comes from the interleave nature but not the better text distribution (MMC4 \vs MMC4-pairs). Blending interleaved and caption data provides a better diversity and downstream accuracy. }
    \label{tab:pt_data_ablate}
\end{table}

\begin{figure*}[t]
    \centering
     \includegraphics[width=\textwidth]{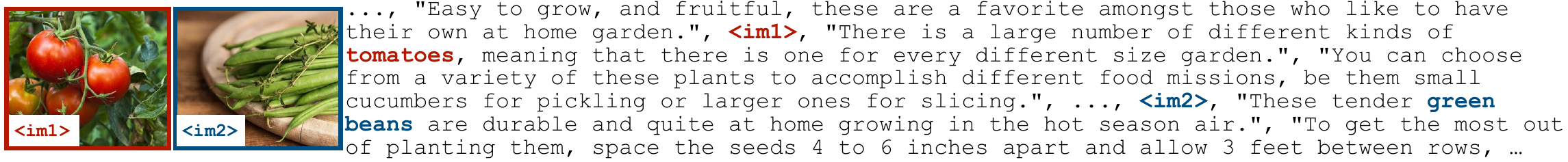}
    \caption{A sample from MMC4~\cite{zhu2023multimodal} dataset consisting of interleaved images and text segments. The images are placed \emph{before} the corresponding text. The text are \emph{weakly conditioned} on images: only colored text can be better inferred with the help of images.
    } 
    \label{fig:mmc4_sample}
\end{figure*}

\begin{figure}[t]
    \centering
     \includegraphics[width=0.45\textwidth]{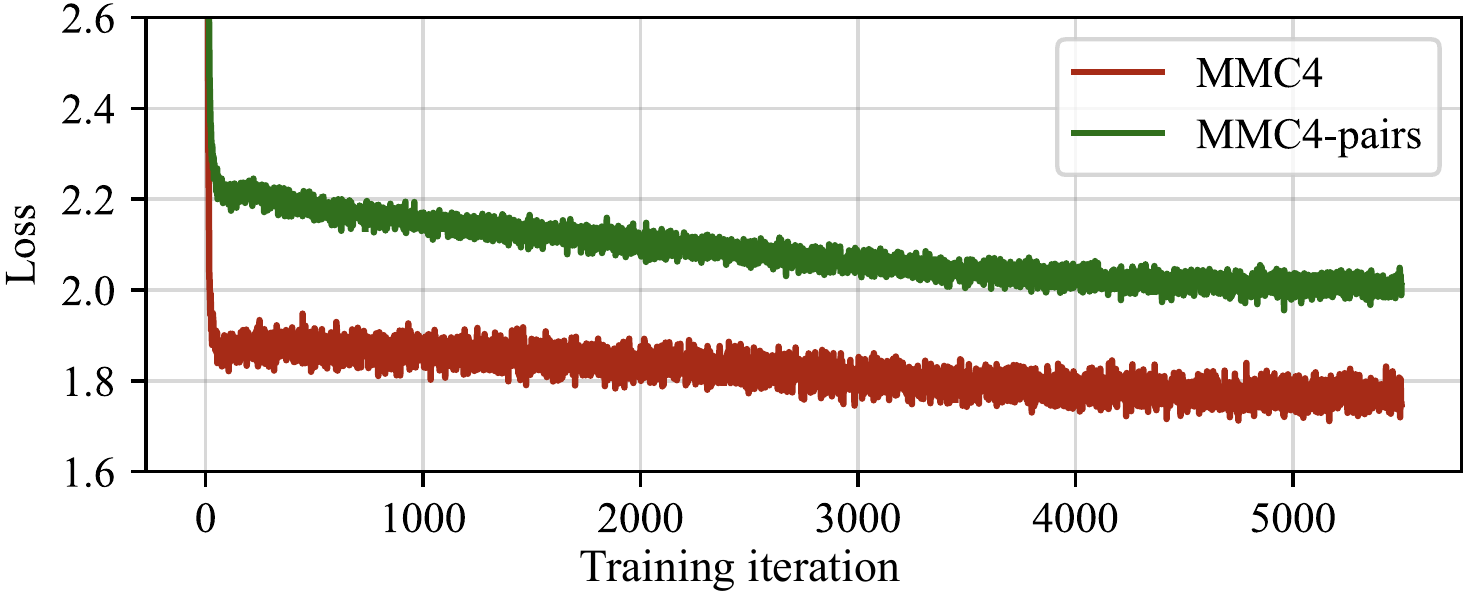}
    \caption{The training loss is lower when pre-training on MMC4 compared to MMC4-pairs (samples broken into image-text pairs), since the text segments provide more information for language modeling.
    } 
    \label{fig:loss_curve}
\end{figure}

Our goal is to ``augment'' the LLM to support visual input, instead of training a model that \emph{only} works well on visual language inputs. Therefore, it is essential to  preserve the text-only capabilities of LLMs. We found that data blending is a key factor, both for pre-training and instruction tuning.

\myparagraph{Pre-training dataset options. }
Most of the VLM pre-training~\cite{liu2023llava, li2023blip, wang2023cogvlm} relies on image-text pairs (\ie, image and captions) due to the wide availability and large diversity (\eg, LAION~\citep{schuhmann2022laion}, COYO~\citep{kakaobrain2022coyo-700m}). On the other hand, interleaved image-text datasets (MMC4~\citep{zhu2023multimodal}, M3W~\citep{alayrac2022flamingo}) follow a more similar distribution compared to the text-only corpus and is found to be important in Flamingo-style model training~\citep{alayrac2022flamingo}. We hypothesize that the interleaved dataset is even \emph{more important} for VLMs when LLM backbone is updated to accommodate the visual input.  
For a better understanding of the two data types, we compare statistics in Table~\ref{tab:pretrain_data}:
COYO suffers from a short text distribution since the accompanying text is taken from alt-text. We subsample the COYO dataset by ranking CLIP similarities and keep only 25M images (a similar size as MMC4-core). 



We follow the same pre-training + SFT process and ablate different pre-training corpus. 
We compare the 0-shot and few-shot visual language accuracy as well as text-only accuracy (MMLU~\cite{mmlu}) in Table~\ref{tab:pt_data_ablate}. Due to space limit, we report the average accuracy over four datasets (as in Table~\ref{tab:training_ablate}). 

\myparagraph{Interleaved data is essential. }
We notice using image-text pairs (\ie, COYO) for pre-training can lead to catastrophic forgetting. The text-only accuracy (MMLU) degrades by $17.2\%$. The visual language accuracy is also much worse compared to MMC4 pre-training. Noticeably, the 4-shot accuracy is even worse than 0-shot, showing the model cannot properly do in-context learning for visual language inputs (probably because it never sees more than one image during pre-training). We hypothesize the catastrophic forgetting is due to the distribution of text-based captions, which are generally very short and concise.

On the contrary, dataset like MMC4 has a much closer distribution compared to text-only corpus (\eg, C4~\cite{raffel2020exploring}). When using the interleaved data for pre-training, the degradation on MMLU is only \textasciitilde $5\%$. The degradation would be even smaller when using a larger base LLM~\citep{driess2023palm}. With proper instruction tuning (Section~\ref{sec:joint_sft}), this degradation can be fully recovered. It also promotes visual in-context learning, leading to a higher 4-shot accuracy compared to 0-shot. 

\myparagraph{Interleave data structure matters, but not the text distribution. }
We further question whether the benefits come from the better text distribution (\eg, longer) or from the interleave nature. To ablate this, we construct a new MMC4 variant by only keeping the images and their corresponding text segments, without considering the interleave nature, denoted as ``MMC4-pairs''.  For example an MMC4 sample may look like:
\begin{align}
\small \nonumber
    \texttt{<txt1>\textbf{<im1>}<txt2><txt3>\textbf{<im2>}<txt4>}
\end{align}
It will be converted into two MMC4-pairs samples\footnote{We followed~\cite{zhu2023multimodal} to match the image and text segments by CLIP scores. }:
\begin{align}
\small \nonumber
    \texttt{\textbf{<im1>}<txt2>, \textbf{<im2>}<txt4>}
\end{align}
However, training on MMC4-pairs does not lead to a satisfactory result: it slightly reduces the degradation on MMLU due to a longer text distribution, but the VLM accuracy is even lower compared to pre-training on COYO; there is also no in-context improvement. We hypothesize the MMC4 samples do not have a very strict image-text correspondence; the image only provides marginal information for text modeling (\ie, most of the information is still from pure text modeling; an example is provided in Figure~\ref{fig:mmc4_sample}). It is also demonstrated by the loss curves in Figure~\ref{fig:loss_curve}, where training on the interleave corpus leads to a much lower loss, indicating the full text segments provides more information. 
Therefore, the interleaved data structure is critical, allowing the model to pick up the image-related information, without over-forcing it to learn unrelated text modeling.

\myparagraph{Data blending improves pre-training. }
Training on image-text pairs only led to a sharp degradation on text-only accuracy (more than $17\%$). Luckily, blending the interleaved corpus and image-text pairs allows us to introduce more diversity in the corpus, while also preventing the severe degradation. Training on MMC4+COYO further boosts the accuracy on visual language benchmarks (the gain is larger when we perform joint SFT, as we will show later (Table~\ref{tab:sft_data_ablate}). 

\subsection{Recover LLM Degradation with Joint SFT}
\label{sec:joint_sft}
\begin{table}[t]
    \setlength{\tabcolsep}{4pt}
    \small
    \centering
    \begin{tabular}{lcccl}
        \toprule
       \multirow{2}{*}{\shortstack{PT\\data}} & \multirow{2}{*}{\shortstack{SFT\\data}} &  \multicolumn{2}{c}{VLM acc. (avg)} &  \multirow{2}{*}{MMLU acc.}    \\  \cmidrule(lr){3-4}  
         &  &  0-shot & 4-shot  \\ \midrule
        \rowcolor{black!10}
         \emph{Llama-2}  & - & - & - & \emph{46.0\%} \\ 
        MMC4 & Visual & 68.7\% & 70.9\%   & 40.7\%  (-5.3\%) \\ 
        MMC4+COYO  & Visual &  69.0\% & 71.3\% & 40.2\% (-5.8\%) \\ \midrule
        \rowcolor{black!10}
        \emph{Llama-2} & \emph{Text} & - & - & \emph{51.2\%} \\  
        MMC4 & Vis.+Text  & 71.0\% & 72.1\% & 51.4\% (\textcolor{mydarkgreen}{+0.2\%}) \\
        MMC4+COYO & Vis.+Text  & \textbf{72.3\%} & \textbf{73.6\%} & 50.9\% (\textcolor{mydarkgreen}{-0.3\%})\\
        \bottomrule
    \end{tabular}
    \caption{Joint SFT (Vis. + Text) not only bridges the degradation of text-only capability (MMLU acc.), but also improves the performance on visual-language tasks (both zero-shot and few-shot).}
    \label{tab:sft_data_ablate}
\end{table}

Despite the interleave data helps maintain the text-only capability, there is still a $5\%$ accuracy drop. 
A potential approach is to maintain the text-only capability would be to add in text-only corpus (the one used in the LLM pre-training). However, such text corpus are usually proprietary even for open-source models; it is also unclear how to subsample the data to match the scale of vision-language corpus. 

Luckily, we found the text-only capabilities are temporarily \emph{hidden}, but not \emph{forgotten}. Adding in text-only data during SFT 
can help bridge the degradation, despite using a much smaller scale compared to the text pre-training corpora (usually trillion scale). 

\myparagraph{Joint supervised fine-tuning. }
The common way for instruction tuning is to fine-tune the model on some visual language datasets (VQA/Caption style~\citep{Dai2023InstructBLIP} or GPT-generated~\citep{liu2023llava}). We found blending in text-only instruction data can simultaneously (i) recover the degradation in \emph{text-only accuracy}, and (ii) improve the \emph{visual language accuracy}. To this end, we also blended in $1$M text-only instruction tuning data sampled from FLAN~\citep{chung2022scaling}, which we termed as \emph{joint SFT}. We provide the comparison in Table~\ref{tab:sft_data_ablate}.



We can see that blending in the text-only SFT data not only bridges the degradation on text-only capability (the MMLU accuracy is on par compared to the original Llama-2 model fine-tuned on the same text-only instruction data), but also improves the visual language capability. We hypothesize that the text-only instruction data improves the model's instruction-following capability, which is also important for visual language tasks. Interestingly, the benefits of blending in COYO data is more significant with joint SFT. We believe that with joint SFT, the model no longer suffers from the text-only degradation when pre-trained with short captions, thus unlocking the full benefits from the better visual diversity.

\begin{table*}[t]
\setlength{\tabcolsep}{3pt}
\centering
\scalebox{0.78}{
\begin{tabular}{l llll | lllll | llllll l }
\toprule
Method & LLM & Res. & PT & IT & VQA$^\text{v2}$ & GQA & VisWiz & SQA$^\text{I}$ & VQA$^\text{T}$ & POPE & MME & MMB & MMB$^\text{CN}$ & SEED & LLaVA$^\text{W}$ & MM-Vet \\
\midrule
BLIP-2~\cite{li2023blip} & Vicuna-13B & 224 & 129M & - & 41.0 & 41 & 19.6 & 61 & 42.5 & 85.3 & 1293.8 & -- & -- & 46.4 & 38.1 & 22.4 \\
InstructBLIP~\cite{Dai2023InstructBLIP} & Vicuna-7B & 224 & 129M & 1.2M & -- & 49.2 & 34.5 & 60.5 & 50.1 & -- & -- & 36 & 23.7 & 53.4 & 60.9 & 26.2 \\
InstructBLIP~\cite{Dai2023InstructBLIP} & Vicuna-13B & 224 & 129M & 1.2M & -- & 49.5 & 33.4 & 63.1 & 50.7 & 78.9 & 1212.8 & -- & -- & -- & 58.2 & 25.6 \\
Shikra~\cite{chen2023shikra} & Vicuna-13B & 224 & 600K & 5.5M & 77.4$^*$ & -- & -- & -- & -- & -- & -- & 58.8 & -- & -- & -- & -- \\
IDEFICS-9B~\cite{idefics} & LLaMA-7B & 224 & 353M & 1M & 50.9 & 38.4 & 35.5 & -- & 25.9 & -- & -- & 48.2 & 25.2 & -- & -- & -- \\
IDEFICS-80B~\cite{idefics} & LLaMA-65B & 224 & 353M & 1M & 60.0 & 45.2 & 36.0 & -- & 30.9 & -- & -- & 54.5 & 38.1 & -- & -- & -- \\
Qwen-VL~\cite{bai2023qwen} & Qwen-7B & 448 & 1.4B & 50M & 78.8$^*$ & 59.3$^*$ & 35.2 & 67.1 & 63.8 & -- & -- & 38.2 & 7.4 & 56.3 & -- & -- \\
Qwen-VL-Chat~\cite{bai2023qwen} & Qwen-7B & 448 & 1.4B & 50M & 78.2$^*$ & 57.5$^*$ & 38.9 & 68.2 & 61.5 & -- & 1487.5 & 60.6 & 56.7 & 58.2 & -- & -- \\
LLaVA-1.5~\cite{liu2023improved} & Vicuna-1.5-7B & 336 & 0.6M & 0.7M & 78.5$^*$ & 62.0$^*$ & 50.0 & 66.8 & 58.2 & \textbf{85.9} & 1510.7 & 64.3 & 58.3 & 58.6 & 63.4 & 30.5 \\ 
LLaVA-1.5~\cite{liu2023improved} & Vicuna-1.5-13B & 336 & 0.6M & 0.7M & \underline{80.0}$^*$ & \textbf{63.3}$^*$ & 53.6 & \underline{71.6} & 61.3 & \textbf{85.9} & 1531.3 & 67.7 & \underline{63.6} & \underline{61.6} & \underline{70.7} & \underline{35.4} \\ \midrule
\rowcolor{black!10}
\methodshort-7B (ours) & Llama-2-7B & 336 & 50M & 1M & 79.9$^*$  & \underline{62.3}$^*$ & \underline{57.8} & 68.2 & \underline{64.4} & \underline{85.5} & \underline{1533.0} & \underline{68.9} & 61.7 & 61.1 & 69.7 & 34.9 \\
\rowcolor{black!10}
\methodshort-13B (ours) & Llama-2-13B & 336 & 50M & 1M & \textbf{80.8}$^*$ & \textbf{63.3}$^*$ & \textbf{60.6} & \textbf{73.7} & \textbf{66.6} & 84.2 & \textbf{1570.1} & \textbf{70.3} & \textbf{64.3} & \textbf{62.8} & \textbf{73.0} & \textbf{38.8} \\
\midrule \midrule
\quad +ShareGPT4V & Llama-2-13B & 336 & 50M & 1M &80.6$^*$ & 63.2$^*$ & 62.4 & 73.1 & 65.3 & 84.8 & 1556.5 & 70.8 & 65.4 & 61.4 & \textcolor{mydarkgreen}{78.4} & \textcolor{mydarkgreen}{45.7} \\
\bottomrule
\end{tabular}
}
\caption{
Comparison with state-of-the-art methods on 12 visual-language benchmarks. Our models consistently outperform LLaVA-1.5 under a head-to-head comparison, using the same prompts and the same base LLM (Vicuna-1.5 is based on Llama-2), showing the effectiveness of visual-language pre-training. We mark the best performance \textbf{bold} and the second-best \underline{underlined}.  
Benchmark names are abbreviated due to space limits. VQA-v2~\cite{goyal2017vqav2}; GQA~\cite{hudson2019gqa}; VisWiz~\cite{gurari2018vizwiz}; SQA$^\text{I}$: ScienceQA-IMG~\cite{lu2022learn}; VQA$^\text{T}$: TextVQA~\cite{singh2019textvqa}; POPE~\cite{li2023pope}; MME~\cite{fu2023mme}; MMB: MMBench~\cite{liu2023mmbench}; MMB$^\text{CN}$: MMBench-Chinese~\cite{liu2023mmbench}; SEED: SEED-Bench~\cite{li2023seed}; LLaVA$^\text{W}$: LLaVA-Bench (In-the-Wild)~\cite{liu2023llava}; MM-Vet~\cite{yu2023mmvet}. $^*$The training images of the datasets are observed during training. 
We also tried adding the ShareGPT4V~\cite{chen2023sharegpt4v} to the SFT blend on top of \methodshort-13B (last row), leading to a significant improvement on LLaVA-Bench and MM-Vet (marked in \textcolor{mydarkgreen}{green}). 
}
\label{tab:sota_comparison}
\end{table*}
\section{Experiments}

\subsection{Scaling up VLM pre-training}
We scale up the training of VLM in the following aspects to form our final model:

\myparagraph{Higher image resolution.} 
Above ablation studies used the OpenAI CLIP-L~\cite{radford2021learning} with 224$\times$224 resolutions as the visual encoder. We now use 336$\times$336 image resolutions to include more visual details for the model, which can help tasks requiring fine-grained details (\eg, TextVQA~\cite{singh2019textvqa}). 


\myparagraph{Larger LLMs.} By default, we used Llama-2~\cite{touvron2023llama2} 7B for ablation study. We also scaled to a larger LLM backbone (\eg, Llama-2~\cite{touvron2023llama2} 13B) to further improve the performance. 

\myparagraph{Pre-training data.} We used both interleaved image-text data and image-text pairs for pre-training (we sample roughly 1:1 image proportions) to improve the data diversity. The total the pre-training corpus contains about 50M images. It is smaller than the billion-scale pre-training data \cite{alayrac2022flamingo, wang2023cogvlm, chen2023pali}, but already demonstrates impressive improvements on downstream tasks. 

\myparagraph{SFT data.} We also include a better SFT data blend from LLaVA-1.5~\cite{liu2023improved}, which is more diverse (\eg, contains reference-based annotations) and has high-quality prompt. The new SFT data blend can significantly improve the downstream evaluation metrics. We include details the Appendix. 

\myparagraph{\emph{Limitations. }} Due to the limited compute budget, we have not been able to further scale up the size of the pre-training corpus to billion-scale, which we leave as future work. Nonethess, pre-training on 50M images already demonstrated significant performance improvement. 

\subsection{Quantitative Evaluation}

\myparagraph{visual language tasks.}
We perform a comprehensive comparison with state-of-the-art models on 12 visual language benchmarks in Table~\ref{tab:sota_comparison}. Compared to existing models (\eg, LLaVA-1.5~\cite{liu2023improved}), our model achieves consistent improvements over most datasets at different model sizes under a head-to-head setting (using the same prompts and base LLM; Vicuna-1.5 is based on Llama-2). 
Remarkably, we 7B model is able to outperform LLaVA-1.5 13B on VisWiz~\cite{gurari2018vizwiz} and TextVQA~\cite{singh2019textvqa} by a large margin thanks to the pre-training. Our 7B model even outperforms the 13B LLaVA model on these datasets. 
Our model also has multi-lingual capability despite the vision-language instruction data is in English, outperforming LLaVA-1.5 on MMBench-Chinese benchmark. 
Our results demonstrates the benefits of vision-language pre-training on downstream tasks, even when using a high-quality instruction tuning dataset~\cite{liu2023improved}. 

\begin{table}[t]
    \setlength{\tabcolsep}{4pt}
    \small
    \centering
    \begin{tabular}{llccc}
        \toprule
       Size & Model & MMLU~\cite{mmlu} & BBH~\cite{bbh} & DROP~\cite{drop} \\ \midrule
        \multirow{4}{*}{7B} &  Llama-2 & 46.0\% & 32.0\% & 31.7\% \\
        & Llama-2+SFT & \textbf{51.8\%} & \textbf{39.3\%} & \textbf{53.1\%} \\
        & Vicuna-1.5 & 49.8\% & 36.9\%  & 29.2\% \\ \cmidrule(lr){2-5}    
        & \methodshort & 50.8\% & 38.5\% & 52.7\% \\ \midrule  
         \multirow{4}{*}{13B} &  Llama-2 & 55.7\% & 37.5\%  & 41.6\% \\
        & Llama-2+SFT &  54.3\% & 43.2\% & 59.2\% \\
        & Vicuna-1.5 & 55.8\% & 38.4\% & 43.6\% \\ \cmidrule(lr){2-5}  
        & \methodshort & \textbf{56.0\%} & \textbf{44.2\%}  & \textbf{63.6\%} \\ 
        \bottomrule
    \end{tabular}
    \caption{\methodshort maintains competitive accuracy on text-only benchmarks. There is a small gap compared to the text-only model under 7B; but the accuracy is even better under 13B.   }
    \label{tab:text_only_acc}
\end{table}

\myparagraph{Text-only performance. }
Our goal is to augment an LLM to support visual inputs. It is essential that the model can retain the text-only capability. 
Therefore, we further evaluate the text-only performance of the models under three benchmarks: MMLU~\cite{mmlu}, BBH~\cite{bbh}, and DROP~\cite{drop} in Table~\ref{tab:text_only_acc}. We did not choose benchmarks like MT-Bench~\cite{zheng2023judging} since text instrution tuning is not the focus of the work. Overall, our model achieves performance as Llama-2 fine-tuned with the same text SFT data: the accuracy of our 7B model is a bit lower, while the 13B is higher. We suspect the smaller model may suffer from a larger text performance degradation during the pre-training, as observed in~\cite{driess2023palm}.

\subsection{Qualitative Evaluation}
Here we study how visual language pre-training enables new capabilities for the model. Part of the image samples are taken from~\cite{yang2023dawn, alayrac2022flamingo, chen2023pali}. 

\begin{figure*}[t]
    \centering
     \includegraphics[width=\textwidth]{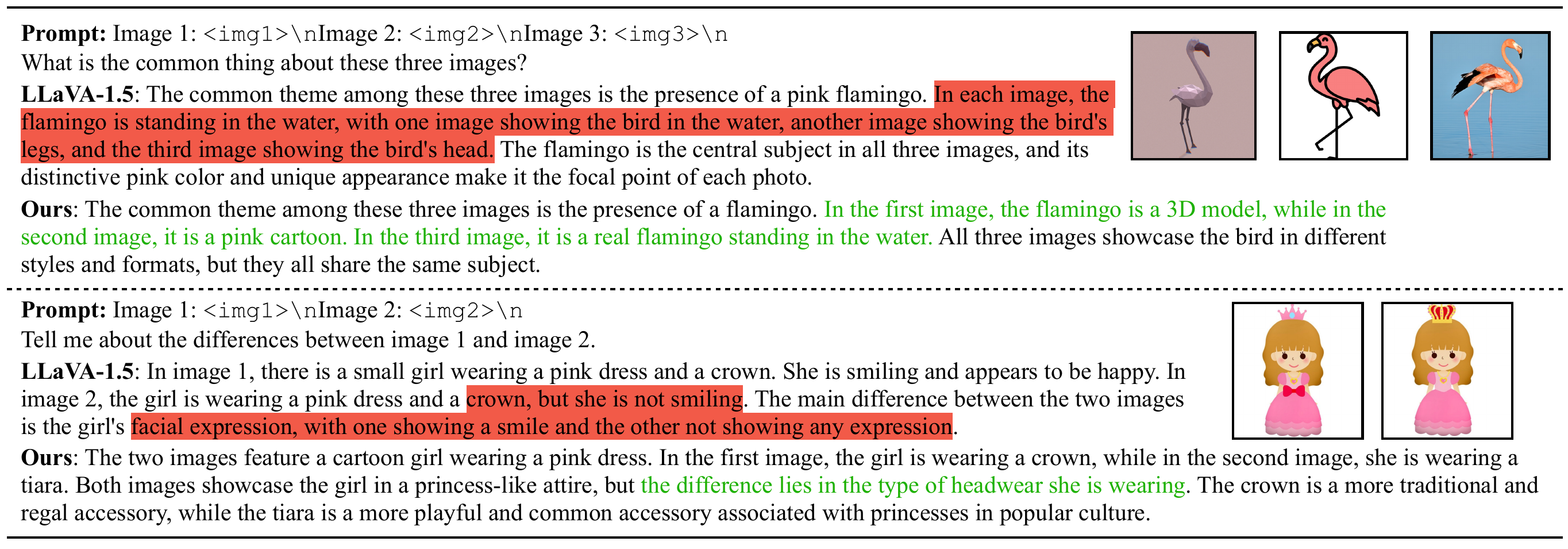}
     \vspace{-10pt}
    \caption{
    Our model \methodshort can reason over multiple images thanks to the pre-training process. The samples are taken from~\cite{alayrac2022flamingo, yang2023dawn}. 
    } 
     \vspace{-8pt}
    \label{fig:multi_image}
\end{figure*}

\myparagraph{Multi-image reasoning. }
Thanks to the pre-training, our model has the ability to reason over multiple images, despite the SFT data is composed of single-image samples. We provide two examples in Figure~\ref{fig:multi_image}. In the first example, our model is able to figure out the common object (\ie, a flamingo) across the three images and the different art styles of each one, while the LLaVA model failed. The LLaVA model hallucinates and cannot distinguish the information from different input images. 
In the second example, our model is able to find one of the differences (\ie, the headwear) out of the two.


\begin{figure}[t]
    \centering
     \includegraphics[width=0.48\textwidth]{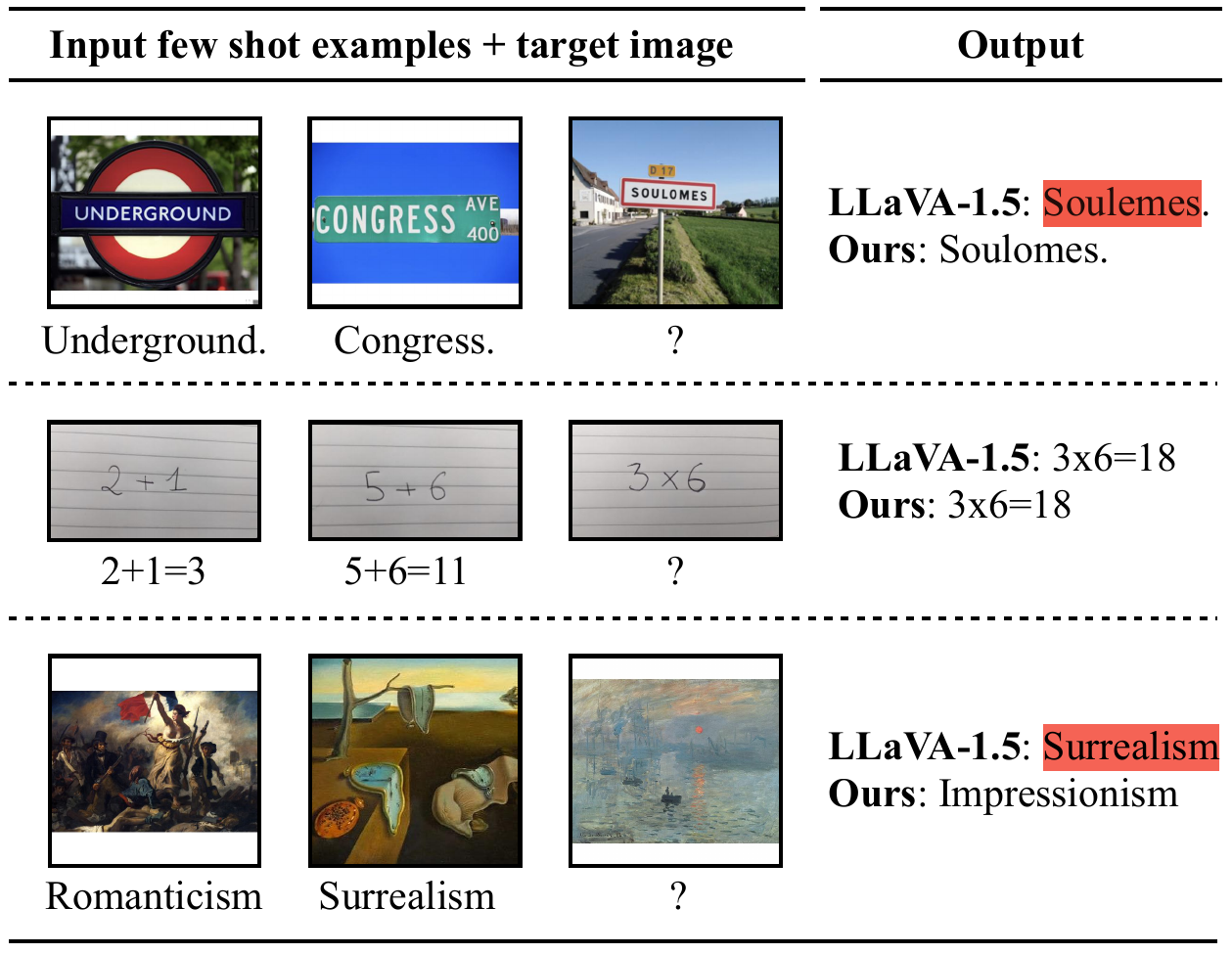}
    \caption{
    \methodshort has better in-context learning capability thanks to interleaved image text pretraining rather than single image-text pairs.We feed two image+text pairs and a third image as the context to prompt the VLM. LLaVA failed the first sample due to limited OCR capability, and failed the third examples by repeating the second sample semantic. 
    } 
    \label{fig:icl_sample}
    \vspace{-10pt}
\end{figure}

\myparagraph{In-context learning.}
In-context learning is an important characteristic of LLMs, allowing people to prompt the LLM with few-shot samples to enable new tasks. We provide in-context learning samples in Figure~\ref{fig:icl_sample}. Interestingly LLaVA-1.5~\cite{liu2023improved} can also perform in-context learning to some extend, despite only being trained on single-image-text-paired samples. We believe the capability is inherited from text-only pre-training of the base LLM. Nonetheless, our model outperforms LLaVA-1.5 for in-context learning: LLaVA-1.5 failed the first sample due to limited OCR capability, and failed the third example by repeating the semantics.




\begin{figure}[t]
    \centering
     \includegraphics[width=0.48\textwidth]{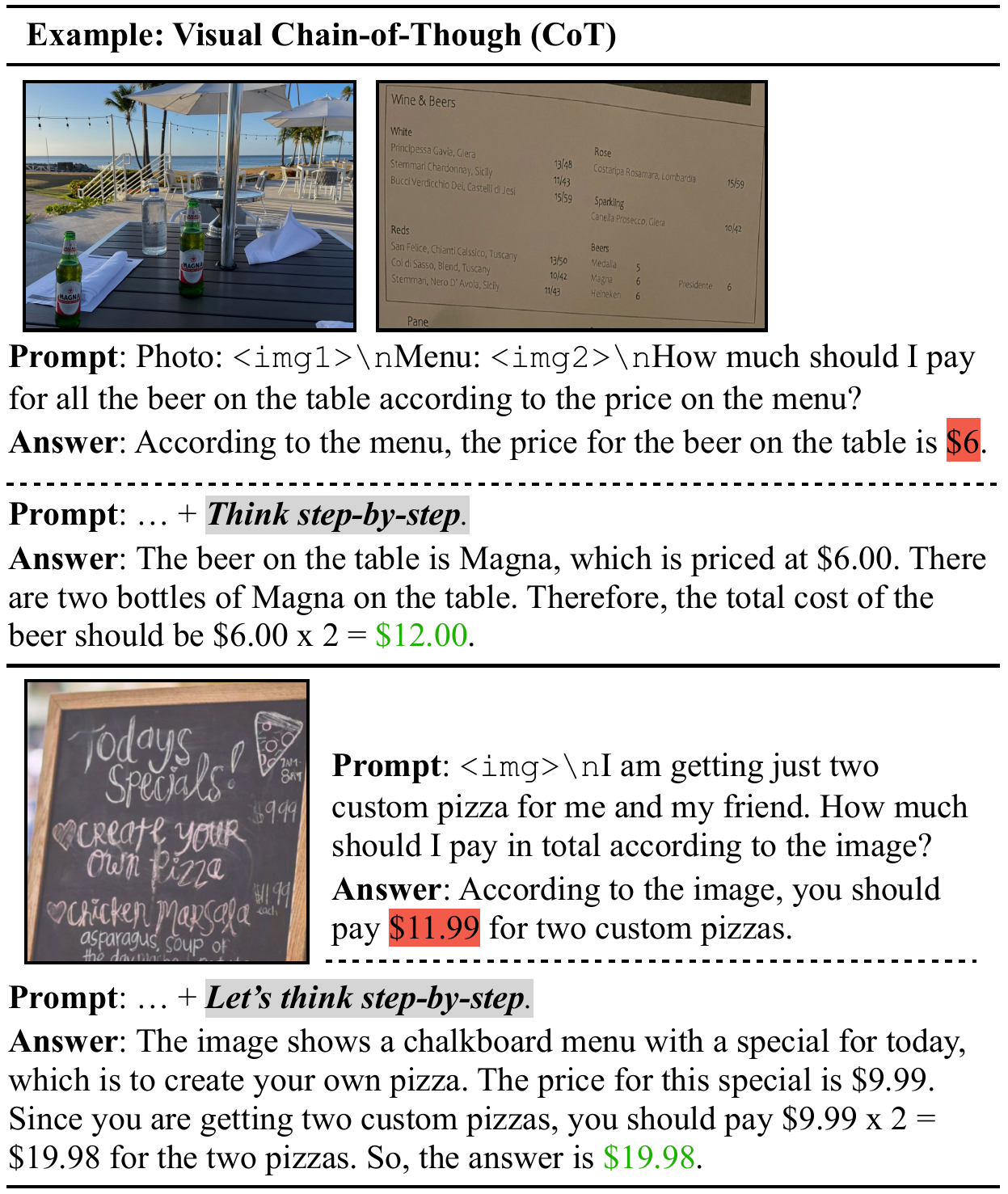}
    \caption{
    Our model is able to perform chain-of-thought reasoning given visual inputs. It is able to generate the correct answer when adding ``Think step-by-step'' to the prompt. \emph{Zoom in} for a better view of the image details. Samples from~\cite{driess2023palm, yang2023dawn}.
    } 
    \label{fig:visual_cot}
\end{figure}

\myparagraph{Visual Chain-of-Thoughts (CoT).}
Our model is able to perform chain-of-thought reasoning given visual inputs. As shown in Figure~\ref{fig:visual_cot}, \methodshort is able to perform complex CoT reasoning over the input images (multi-image or single-image) when adding ``Think step-by-step'' to the end of the prompt. 
We believe the CoT capability is inherited from text-only SFT, despite there is no such samples from the visual language instruction data. 

\myparagraph{Better world knowledge. }
Since our model is pre-trained on a large-scale corpus, it has better understanding of world knowledge. We perform a case study by prompting the model to recognize the locations of some famous landmarks (please see supplementary due to space limits). \methodshort can correctly recognize 4 out of the 4 samples, while LLaVA-1.5 only gets 2 out of the 4, demonstrating the effectiveness of the pre-training. Samples are taken from~\cite{yang2023dawn}. 



\subsection{Other Learnings.}

\begin{table}[t]
    \setlength{\tabcolsep}{3pt}
    \small
    \centering
    \begin{tabular}{llcccc}
        \toprule
        \multirow{2}{*}{Resolution} & \multirow{2}{*}{Projector} & \multirow{2}{*}{\#Tokens} & \multicolumn{3}{c}{Evaluation $\uparrow$} \\  \cmidrule(lr){4-6}  
        & & & OKVQA & TextVQA & COCO \\ \midrule
        224 & linear & 256 & 49.9\% & \textcolor{mydarkred}{41.6\%} & 116.0 \\ 
        336 & linear & 576 & 49.7\% & \textcolor{mydarkgreen}{49.8\%} & 117.7 \\ \midrule
        336 & downsample & {144} & 49.3\% & 45.6\% & 115.7 \\
        \bottomrule
    \end{tabular}
    \caption{ Improving the image resolution from 224 to 336 can significantly improve TextVQA accuracy. The raw resolution matters more than \#tokens; high-resolution with token downsampling works better than low-resolution. We report accuracy for OKVQA and TextVQA, and CIDEr for COCO. 
    Note: the evaluation protocol is different from Table~\ref{tab:sota_comparison} and can only be compared within the table.  }
    \vspace{-8pt}
    \label{tab:resolution_study}
\end{table}

\myparagraph{Image resolution matters, not \#tokens. }
We chose an image resolution of $336^2$ since it provides more fine-grained details compared to $224^2$, leading to improved accuracy on tasks like TextVQA~\cite{singh2019textvqa}. As shown in Table~\ref{tab:resolution_study}, increasing the resolution from 224 to 336 can improve the TextVQA accuracy from $41.6\%$ to $49.8\%$.  However, a higher resolution leads to more tokens per image ($336^2$ corresponds to 576 tokens/image) and a higher computational cost. It also limits the number of demonstrations for in-context learning. 

Luckily, we find that the raw resolution matters more than the \#visual tokens/image. We can use different projector designs to compress the visual tokens. Here we try a ``downsample'' projector, which
simply concatenates every $2\times2$ tokens into a single one and use a linear layer to fuse the information. It reduces the \#tokens to 144 under the 336 resolution,  
that is even smaller than the 224+linear setup. Nonetheless, the TextVQA accuracy is higher ($\sim$$46\%$ \vs $41.6\%$), despite still $3\%$ worse compared to 336+linear setup, showing a large redundancy in the image tokens. The gap on other datasets such as OKVQA and COCO is smaller since they usually require higher-level semantics. 

In our main results, we did not use any token compression methods to provide the best accuracy despite this encouraging observation, and leave it to future work. 

\myparagraph{Comparison to frozen LLMs with visual experts. }
Another interesting method for retaining the text capabilities of LLMs during the pre-training is to freeze the base LLM and add an extra visual expert to process the visual tokens~\cite{wang2023cogvlm}. The definition of expert is similar to MoE frameworks, but with a manual routing mechnism according to token types. Since the base LLM is frozen, the model fully retains the original functionality for text-only inputs during pre-training. However, we find that directly fine-tuning the LLM during visual language pre-training still leads to a better VLM accuracy and in-context learning capability (Table~\ref{tab:compare_vis_exp}). Adding an extra visual expert also leads to near 2$\times$ model size increase, which is not friendly for edge deployment. Therefore, we chose to directly fine-tune the base LLM. 

\begin{table}[t]
    \small
    \centering
    \begin{tabular}{cccl}
        \toprule
        \multirow{2}{*}{} & \multirow{2}{*}{\#Param} & \multicolumn{2}{c}{VLM acc. (avg)}\\  \cmidrule(lr){3-4}  
         & &  0-shot & 4-shot  \\ \midrule
        Visual Expert~\cite{wang2023cogvlm} & 1.9$\times$ & 67.0\% & 64.8\% \\
        Fine-tune & \textbf{1$\times$} & \textbf{71.0\%} & \textbf{72.1\%} \\
        \bottomrule
    \end{tabular}
    \vspace{-5pt}
    \caption{Directly fine-tuning the LLM during pre-training leads to better VLM accuracy and in-context learning capabilities. It also enjoys a smaller model size. Both settings are pre-trained on the MMC4-core dataset~\cite{zhu2023multimodal}. 
    \vspace{-15pt}
    }
    \label{tab:compare_vis_exp}
\end{table}

\myparagraph{Comparison to PEFT/LoRA.} 
In addition to visual experts, we also provide extra results when performing LoRA tuning with rank $64$ (7B model) in Table~\ref{tab:compare_lora}. Fine-tuning LLM outperforms LoRA tuning by a large margin. 

\begin{table}[h]
    \setlength{\tabcolsep}{2pt}
    \small
    \centering
    \begin{tabular}{lccccccc}
        \toprule
         & VQA$^\text{v2}$ & GQA & VisWiz & VQA$^\text{T}$ & MMB &  LLaVA$^\text{W}$ \\ \midrule
        LoRA ($r$=64) & 69.4 & 54.3 & 48.4 & 50.0 & 60.3 & 51.2 \\
        Fine-tune (ours) & \textbf{79.9} & \textbf{62.3} & \textbf{57.8} &  \textbf{64.4} & \textbf{68.9} & \textbf{69.7} \\
        \bottomrule
    \end{tabular}
    \vspace{-5pt}
    \caption{Fine-tuning LLM consistently outperforms LoRA tuning. 
    \vspace{-15pt}
    }
    \label{tab:compare_lora}
\end{table}

\myparagraph{Reformatting the interleaved structure.} For additional insights we also reformatted the MMC4 dataset to be \texttt{<im1><im2><txt1><txt2>} instead of the \texttt{<im1><txt1><im2><txt2>} and evaluated the model under the setting in Table~\ref{tab:training_ablate}. We observed that the reformatted MMC4 degrades the average 0-shot accuracy (on the 4 benchmarks) by \emph{4.4\%}, and degrades the average 4-shot accuracy by \emph{37.5\%}. The disorder breaks in-context learning capability, showing the importance of interleaved data. 
\section{Related Work}  

\myparagraph{Large language models (LLMs).}
LLMs based on Transformers~\cite{vaswani2017attention} have fundamentally changed the language processing field. They are achieving increasing capabilities by \emph{scaling up} the model size and the pre-training corpus~\cite{gpt3, GPT4, smith2022using, du2022glam, fedus2022switch, rae2021scaling, hoffmann2022training, chowdhery2022palm, dai2019transformer}.
It is believed that most the capability of the LLM is obtained from the large-scale \emph{pre-training} process, which are later unlocked through instruction tuning~\cite{ouyang2022training, OpenAI_ChatGPT, chung2022scaling}. There is a growing effort from the open-source community to build a strong base LLM~\cite{opt, touvron2023llama, touvron2023llama2}, the conversational variants~\cite{alpaca, vicuna2023} and the parameter-efficient finetuned versions of large LLMs~\cite{luo2024cheap,zhang2023llama}. In this work, we start with the base Llama-2 model~\cite{touvron2023llama2}. 
\myparagraph{Visual language models (VLMs).}
VLMs are LLMs augmented with visual inputs to provide a unified interface for visual language tasks. There are two main designs for VLMs: 1. cross-attention based, where the LLM is frozen while the visual information is fused into intermediate embeddings with a cross-attention mechanism~\cite{alayrac2022flamingo, openflamingo}; 2. auto-regressive based, where the visual input is tokenized and fed to the LLM alongside text tokens~\cite{liu2023llava, driess2023palm, chen2023pali, li2023blip, fuyu, zhu2023minigpt, ye2023mplug, qwen, aiello2023jointly}. The latter is a natural extension by treating visual inputs as a foreign language. 
VLMs are also instruction-tuned so that they can better follow human instructions or perform conversations~\cite{Dai2023InstructBLIP, liu2023llava, sun2023aligning}. 
In this work, we study the pre-training process of the auto-regressive VLMs due to their flexibility when handling multi-modal inputs. 

Following text-only LLMs, people also study different training recipes for VLMs. Some work freezes the LLM and train auxiliary components~\cite{alayrac2022flamingo, li2022blip, li2023blip, wang2023cogvlm}, others fine-tune the LLM to enable visual capabilities~\cite{chen2023pali, driess2023palm, zhao2023mmicl}. There is also usage of different data corpora, including image-text pairs~\cite{chen2023pali, li2022blip, driess2023palm, liu2023llava}, interleaved datasets~\cite{openflamingo}, video-text pairs~\cite{maaz2023video}, visual-grounded annotations~\cite{peng2023kosmos, liu2023improved}, \etc. 
In this work, we provide a holistic ablation of different design choices for the pre-training stage.





\section{Conclusion}
This paper has explored effective pretraining design options to augment LLMs towards vision tasks. Leveraging full strength of LLM learning, interleaved-nature of image-text data, and careful text data re-blending, \methodshort has surpassed state-of-the-art methods for vision tasks while preserving text-only capabilities. \methodshort has also depicted strong reasoning capability for multi-image analysis, in-context learning and zero/few-shot tasks. We hope our paper can help spur further research on VLM pretraining and collection of cross-modality datasets.


\section*{Acknowledgements}
We would like to thank Bryan Catanzaro for fruitful discussions. We also appreciate the help from Zhuolin Yang, Guilin Liu, Lukas Voegtle, Philipp Fischer, Karan Sapra and Timo Roman on dataset preparation and feedback. 


{\small
\bibliographystyle{ieeenat_fullname}
\bibliography{main}
}

\clearpage
\appendix

\section{SFT Blend for Ablation Study}
\label{sec:supp_sft_blend}

We used an in-house data blend for supervised fine-tuning/instruction tuning during the ablation study. We followed~\cite{Dai2023InstructBLIP} to build the FLAN-style instructions from the training set of 18 visual language datasets, as shown in Table~\ref{tab:sft_blend}. 
We may see that most of the datasets are in a VQA format.
For the final model, we also blend in the LLaVA-1.5 SFT dataset~\cite{liu2023improved}, which has better quality and diversity (for example, it contains visual reference data like RefCOCO~\cite{lin2014mscoco, mao2016generation}). 

\begin{table}[h]
    \setlength{\tabcolsep}{3.5pt}
    \small
    \centering
    \begin{tabular}{l|l}
        \toprule
        \textbf{Categories} & \textbf{Datasets} \\ \midrule
        Captioning & Image Paragraph Captioning, MSR-VTT, TextCaps \\
        Reasoning & CLEVR, NLVR, VisualMRC \\
        Translation & Multi30k \\
        VQA & \parbox[t]{6cm}{ActivityNet-QA, DocVQA, GQA, iVQA, MSRVTT-QA, MSVD-QA, OCR-VQA, ST-VQA, ViQuAE, VQAv2, Visual Dialog} \\
        \bottomrule
    \end{tabular}
    \caption{ The SFT blend we used during the ablation study. }
    \label{tab:sft_blend}
\end{table}

\section{Training Cost}

We perform training on 16 A100 GPU nodes, each node has 8 GPUs. The training hours for each stage of the 7B model are: 
projector initialization: 4 hours; 
visual language pre-training: 30 hours; 
visual instruction-tuning: 6 hours. 
The training corresponds to a total of 5.1k GPU hours. Most of the computation is spent on the pre-training stage. 


We have not performed training throughput optimizations like sample packing~\cite{kosec2021packing} or sample length clustering. We believe we can reduce at least 30\% of the training time with proper optimization. We also notice that the training time is much longer as we used a high image resolution of 336$\times$336 (corresponding to 576 tokens/image). We should be able to reduce the training time by more than 50\% by using lower-resolution images for pre-training (\eg, 224$\times$224) and scale up the resolution at the later stage of the training~\cite{chen2023pali}, which we leave to future work. 

\section{Varying LLMs}

For extra insights we here present the results with Vicuna-1.5-7B as the new LLM backbone to verify if our pretraining conclusions scale across LLMs. Firstly, we reproduce the training recipe study (originally in Table~\ref{tab:training_ablate}) and report the average 0/4-shot accuracy in Table~\ref{tab:vicuna} (left). We observed the same conclusion that updating the LLM is important in the pre-training stage. Secondly, we also provide the results in the setting of Table~\ref{tab:sota_comparison} (as right of Table~\ref{tab:vicuna}). The two backbones achieved similar accuracy on the benchmarks. Overall, our conclusions are general across LLM backbones. 

\begin{table}[h]
    \setlength{\tabcolsep}{2pt}
    \small
    \centering
    \begin{tabular}{cccc}
        \toprule
        PreT & SFT & 0-shot & 4-shot \\ \midrule
        \ding{55} & \ding{51} & 62.6 & 59.2 \\
        \ding{51} & \ding{51} & \textbf{69.1} & \textbf{72.8}\\
        \bottomrule
    \end{tabular}
    \quad
    \begin{tabular}{lccc}
        \toprule
          & VQA$^\text{v2}$ & GQA & VisWiz \\ \midrule
       Llama-2 7B & 79.9 & 62.3 & 57.8 \\
       Vicuna 7B & 79.3 & 62.3 &	58.7 \\
        \bottomrule
    \end{tabular}
    \vspace{-5pt}
    \caption{Ablation and final performance with Vicuna-1.5-7B. 
    \vspace{-12pt}
    }
    \label{tab:vicuna}
\end{table}

\section{Details on COYO Subsampling}

We were able to download 25M out of 30M images for the MMC4-core dataset~\cite{zhu2023multimodal}. The COYO-700M dataset~\cite{kakaobrain2022coyo-700m} contains about 700M images. To maintain a similar dataset size, we subsample 25M images from the COYO-700M dataset. Specifically, we sort all the samples based on the CLIP similarity between images and captions and keep the 25M images with the highest similarities. Samples with a high CLIP similarity usually have better image-caption correspondence.

\section{More Qualitative Samples}

Here we provide more qualitative samples that we were not able to include in the main paper due to space limits. Many of the image samples are taken from~\cite{yang2023dawn, alayrac2022flamingo}. 


\begin{figure}[h]
    \centering
     \includegraphics[width=0.4\textwidth]{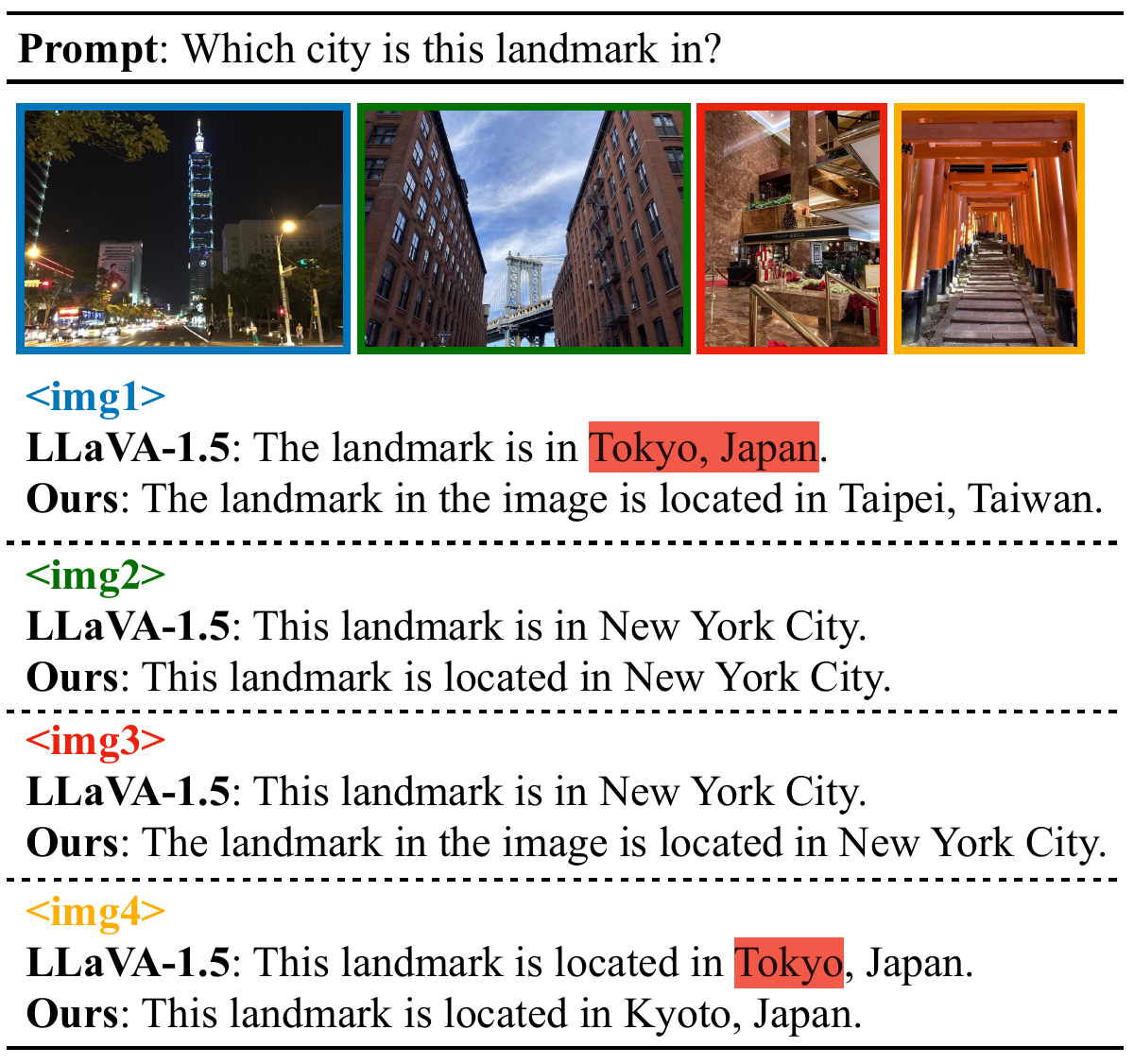}
    \caption{Landmark city recognition. Visual-language pre-training gives the model better world knowledge. It reduces the bias towards answering ``Tokyo'' compared to LLaVA-1.5~\cite{liu2023improved}. We mark the wrong responses in \textcolor{mydarkred}{red}. 
Samples are taken from~\cite{yang2023dawn}.
    } 
    \label{fig:landmark}
\end{figure}

\paragraph{Better world knowledge. }
Pre-training on a large-scale corpus allows the model to have better visual-related world knowledge. Here we take four landmark images from~\cite{yang2023dawn} (without curation) and ask the model which city is the landmark located in (Figure~\ref{fig:landmark}). \methodshort can correctly recognize 4 out of the 4 samples, while LLaVA-1.5 only gets 2 out of the 4, with an output bias on more common cities like Tokyo and New York.

\begin{figure}[t]
  \begin{minipage}{0.99\linewidth}
  \small
\centering
\scalebox{0.8}{
\begin{tabular}{p{1.6cm} p{8.0cm} }
\toprule
 \multicolumn{2}{l}{\bf Visual input example: Understanding visual reference}  \\
\midrule
&  \includegraphics[height=3.5cm]{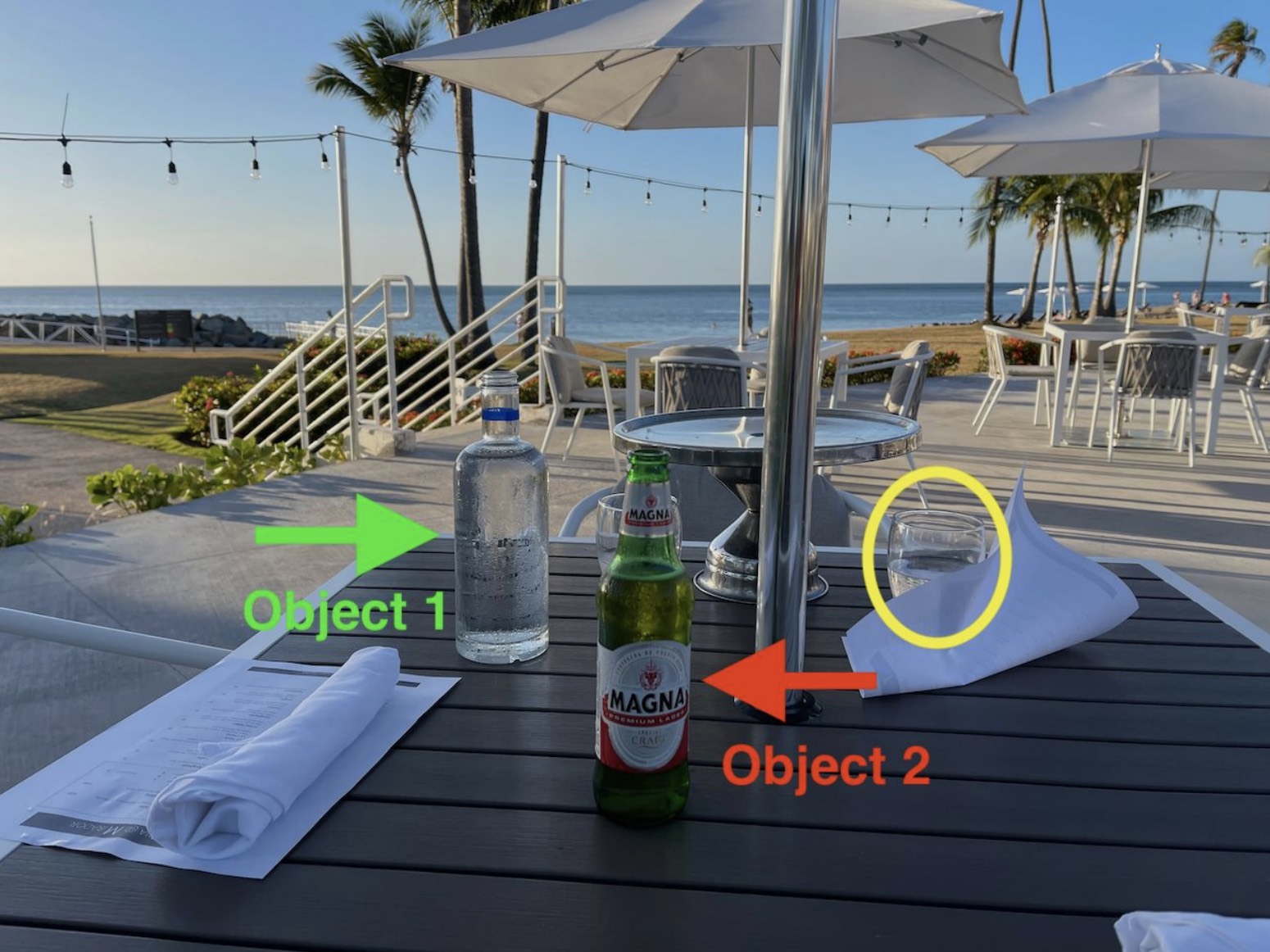} \\

\textbf{User} & What is in the circled glass? Object 1 or 2? Describe what is Object 1 and object 2. Then check what is in the circled glass. \\
\midrule
\textbf{GPT-4V} & Object 1 is a glass bottle of water. It is clear and has a blue label on it.
Object 2 is a glass bottle of Magna beer. It is
green and has a red and white label on it.
In the circled glass, it appears to be a clear
liquid, likely water. So, it is likely that the
contents of the circled glass are from Object 1,
the glass bottle of water. \\
\midrule
\textbf{LLaVA-1.5} & Object 1 is a bottle of water, and Object 2 is a bottle of beer. The circled glass contains the \hl{beer}. \\
\midrule
\textbf{Ours} & Object 1 is a bottle of water, and Object 2 is a bottle of beer. The circled glass contains Object 1, which is a bottle of water. \\
\bottomrule
\end{tabular}
}
\caption{Our model can understand visual reference overlaid on images and perform reasoning. 
}
\label{tab:demo_bottles}  
  \end{minipage}
\end{figure}

\paragraph{Visual reference understanding.}
Our model can understand visual reference overlaid on images and perform reasoning. We provide a sample of visual reference reasoning in Figure~\ref{tab:demo_bottles} (from~\cite{yang2023dawn}). \methodshort is able to correctly figure out what is in the circled glass, while LLaVA-1.5 failed.

\begin{figure}[t]
  \begin{minipage}{0.99\linewidth}
  \small
\centering
\scalebox{0.8}{
\begin{tabular}{p{1.6cm} p{8.0cm} }
\toprule
 \multicolumn{2}{l}{\bf Visual input example: Visual logic reasoning}  \\
\midrule
&  \includegraphics[height=3.5cm]{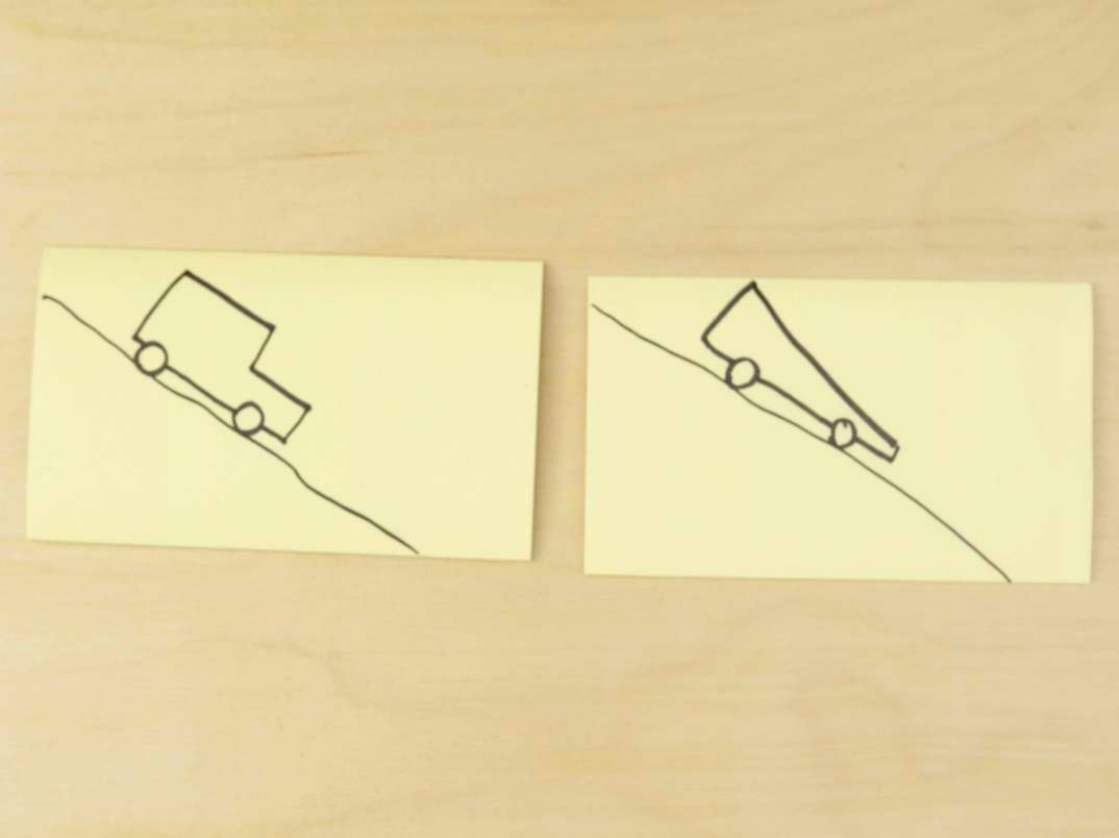} \\

\textbf{User} & Which of these cars is more aerodynamic? The one on the left or the right? Explain why, using specific visual details. \\
\midrule
\textbf{Gemini} & The car on the right is more aerodynamic. It has a lower profile and a more streamlined shape. The car on the left has a higher profile and a more boxy shape, which makes it less aerodynamic. \\
\midrule
\textbf{LLaVA-1.5} & The car on the right is more aerodynamic than the one on the left. This is evident from the fact that \hl{the car on the right is shown going down a hill, while the car on the left is shown going up a hill}. When a car is going {downhill}, it experiences less air resistance, which allows it to maintain higher speeds with less energy expenditure. In contrast, when a car is going {uphill}, it encounters more air resistance, making it more challenging to maintain speed and requiring more energy to overcome the resistance. Therefore, the car on the right is more aerodynamic due to the downhill slope, which allows it to travel more efficiently.
 \\
\midrule
\textbf{Ours} & The car on the right is more aerodynamic than the one on the left. This is because the car on the right has a more streamlined shape, which reduces air resistance and drag. The car on the left has a more traditional, boxy shape, which increases air resistance and drag. The aerodynamic car on the right would be more efficient and faster, while the car on the left would be slower and less efficient. \\
\bottomrule
\end{tabular}
}
\caption{Our model can understand visual details on images and perform logical reasoning. 
}
\label{tab:gemini}  
  \end{minipage}
\end{figure}

\paragraph{More logical reasoning samples. } We check VILA on the most recent samples from Gemini's release~\cite{gemini} in Figure~\ref{tab:gemini}. \methodshort is able to understand the logic utilizing visual features in detail, whereas the LLAVA-1.5 cannot yield reasonable responses.

\begin{figure}[h]
    \centering
     \includegraphics[width=0.5\textwidth]{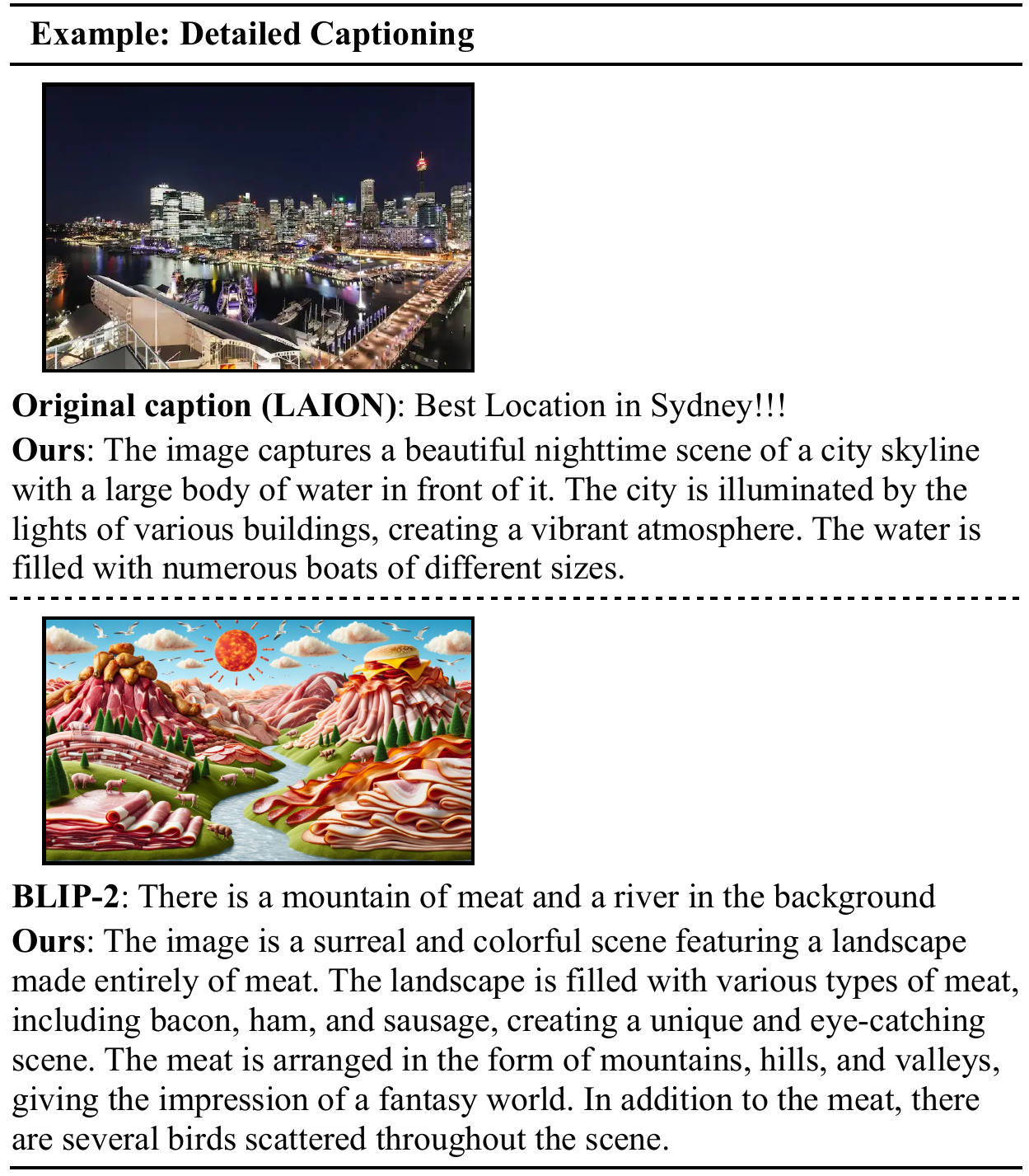}
    \caption{\methodshort can provide detailed captions. The raw captions in datasets like LAION~\cite{schuhmann2022laion} can be noisy and irrelevant. \methodshort can generate meaningful captions with more details compared to BLIP-2~\cite{li2023blip}. 
    The results are obtained by prompting the model with ``Describe the image in detail.''. 
    } 
    \label{fig:supp_detailed_caption}
\end{figure}

\paragraph{Using \methodshort for detailed captioning. }
People have been using datasets like LAION~\cite{schuhmann2022laion} to train text-to-image generative models~\cite{rombach2022high, ramesh2022hierarchical}. The quality of the image-text pairs can significantly affect the performance of the trained model. Some captions in the training datasets are quite noisy: they are either not quite related to the images or are too abbreviated and contain limited details. We show that we can use VLM models to generate high-quality and detailed captioning (Figure~\ref{fig:supp_detailed_caption}). We use a simple prompt ``Describe the image in detail.'' to generate the captions. \methodshort can generate more related descriptions compared to the original caption (sample 1) and provide more details compared to previous models like BLIP-2~\cite{li2023blip} (sample 2).

\begin{figure*}[h]
    \centering
     \includegraphics[width=0.95\textwidth]{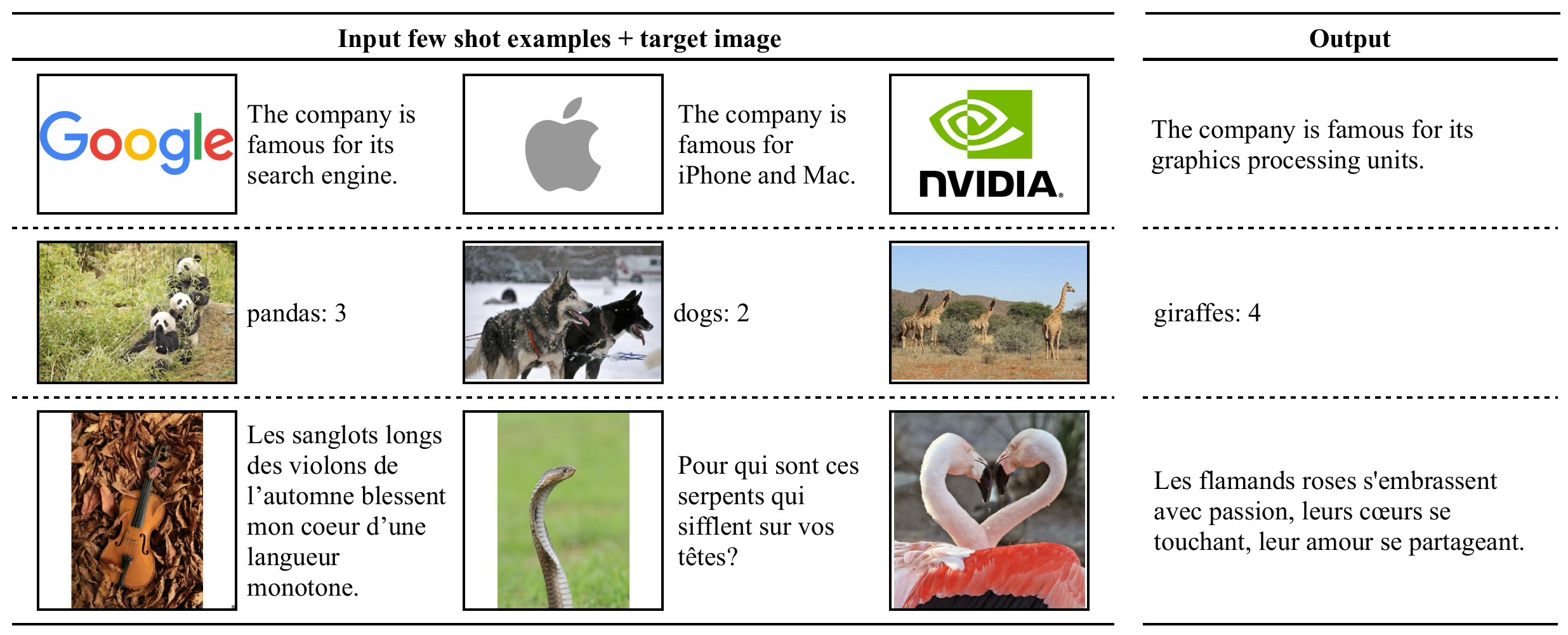}
    \caption{In-context learning samples on company knowledge, object counting, and French poem. The predictions are from \methodshort-13B. 
    } 
    \label{fig:supp_icl_samples}
\end{figure*}

\paragraph{More in-context learning samples. } We provide more in-context learning samples in Figure~\ref{fig:supp_icl_samples}, including company knowledge, object counting, and French poems. \methodshort demonstrates strong in-context learning capabilities under various demonstrations.

\begin{figure*}[h]
    \centering
     \includegraphics[width=0.95\textwidth]{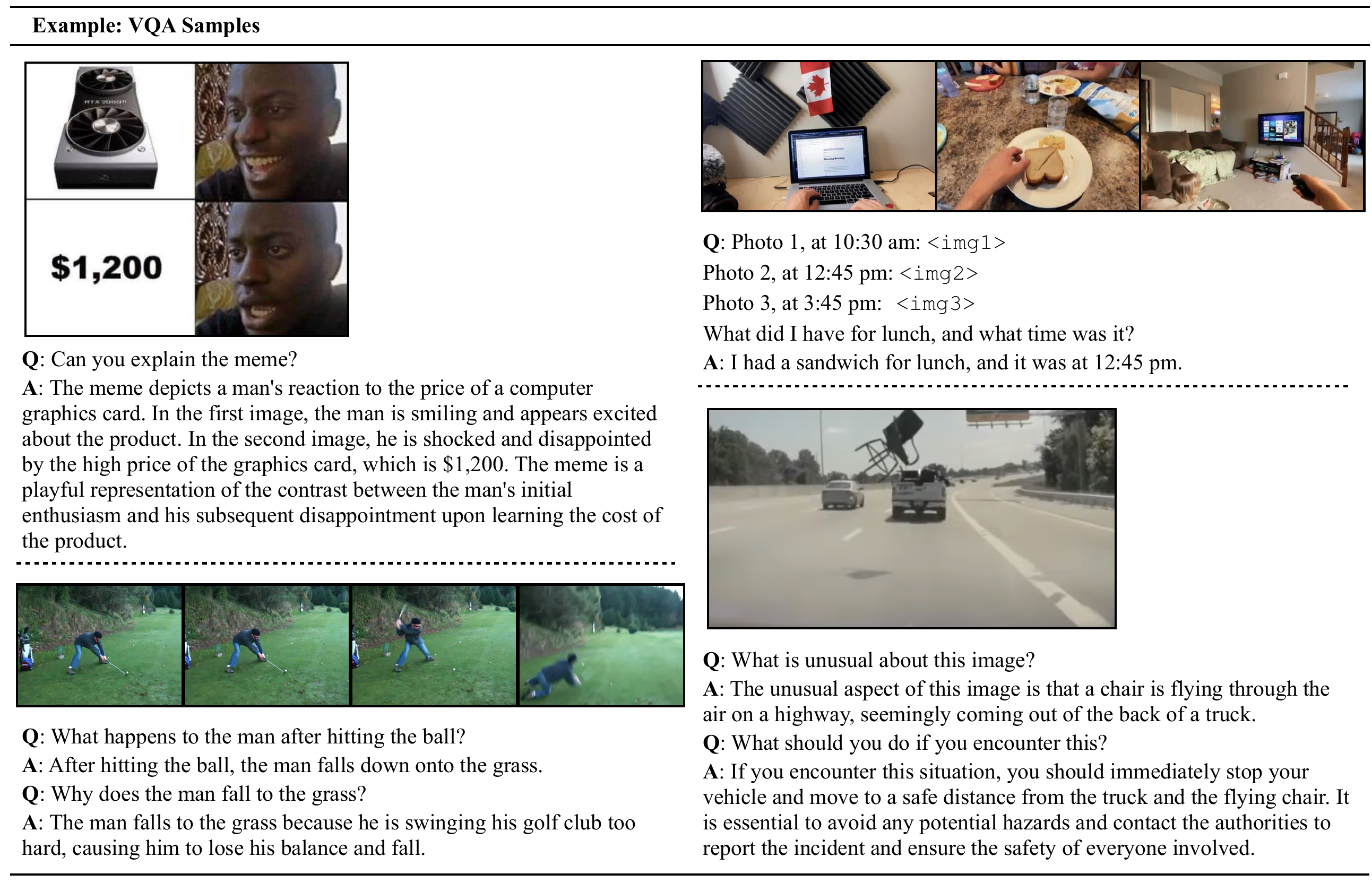}
    \caption{VQA samples. \methodshort is able to understand memes, reason on multiple images or video frames, and provide help on corner cases in autonomous driving. The answers are from \methodshort-13B. 
    } 
    \label{fig:supp_vqa_samples}
\end{figure*}

\paragraph{More VQA samples. } We provide more VQA samples in Figure~\ref{fig:supp_vqa_samples}. \methodshort is able to understand memes, reason on multiple images or video frames, and provide help on corner cases in autonomous driving.

\clearpage

\end{document}